\PassOptionsToPackage{numbers, sort&compress}{natbib}
\documentclass{article}

\usepackage{arxiv}

\usepackage[utf8]{inputenc} 
\usepackage[T1]{fontenc}    
\usepackage{hyperref}       
\usepackage{url}            
\usepackage{booktabs}       
\usepackage{amsfonts}       
\usepackage{nicefrac}       
\usepackage{microtype}      
\usepackage{setspace}
\usepackage{graphicx}
\usepackage{natbib}
\usepackage{doi}
\usepackage{algorithm}
\usepackage{algpseudocode} 
\usepackage[table]{xcolor} 
\usepackage{colortbl} 
\usepackage{multirow}
\usepackage{amsmath}
\usepackage{amssymb}
\usepackage{amsfonts}
\usepackage{amsthm} 
\newtheorem{definition}{Definition}
\usepackage{soul}
\usepackage[table]{xcolor}
\usepackage{wrapfig}
\definecolor{colorA}{rgb}{0.85, 0.9, 1} 
\definecolor{colorB}{rgb}{0.9, 1, 0.9} 
\definecolor{colorC}{rgb}{1, 0.9, 0.8}

\newcommand{\violetcell}[1]{\cellcolor{violet!8}#1}

\usepackage[capitalize]{cleveref} 
\crefname{supp}{Supp.}{Supp.}

\usepackage{xspace}
\makeatletter
\DeclareRobustCommand\onedot{\futurelet\@let@token\@onedot}
\def\@onedot{\ifx\@let@token.\else.\null\fi\xspace}

\def\eg{\emph{e.g}\onedot}

\def\etal{\emph{et al}\onedot}
\makeatother
\newcommand{\thirdorder}[1]{\vspace{1mm}\noindent\textbf{#1} }

\title{MotionBits: Video Segmentation through Motion-Level Analysis of Rigid Bodies}

\author{{Howard H. Qian} \\
	Department of Computer Science\\
	Rice University\\
	\And
	{Kejia Ren} \\
	Department of Computer Science\\
	Rice University\\
	\And
	{Yu Xiang} \\
	Department of Computer Science\\
	University of Texas at Dallas\\
    \And
	{Vicente Ordonez} \\
	Department of Computer Science\\
	Rice University\\
    \And
	{Kaiyu Hang} \\
	Department of Computer Science\\
	Rice University\\
}

\date{}

\renewcommand{\shorttitle}{MotionBits: Video Segmentation through Motion-Level Analysis of Rigid Bodies}

\begin{document}
\maketitle

\begin{abstract}
  Rigid bodies constitute the smallest manipulable elements in the real world, and understanding how they physically interact is fundamental to embodied reasoning and robotic manipulation. Thus, accurate detection, segmentation, and tracking of moving rigid bodies is essential for enabling reasoning modules to interpret and act in diverse environments. However, current segmentation models trained on semantic grouping are limited in their ability to provide meaningful interaction-level cues for completing embodied tasks. To address this gap, we introduce MotionBit, a novel concept that, unlike prior formulations, defines the smallest unit in motion-based segmentation through kinematic spatial twist equivalence, independent of semantics. In this paper, we contribute (1) the MotionBit concept and definition, (2) a hand-labeled benchmark, called MoRiBo, for evaluating moving rigid-body segmentation across robotic manipulation and human-in-the-wild videos, and (3) a learning-free graph-based MotionBits segmentation method that outperforms state-of-the-art embodied perception methods by 37.3\% in macro-averaged mIoU on the MoRiBo benchmark. Finally, we demonstrate the effectiveness of MotionBits segmentation for downstream embodied reasoning and manipulation tasks, highlighting its importance as a fundamental primitive for understanding physical interactions.
\end{abstract}

\begin{figure}[t]
    \centering
    \includegraphics[width=0.85\linewidth]{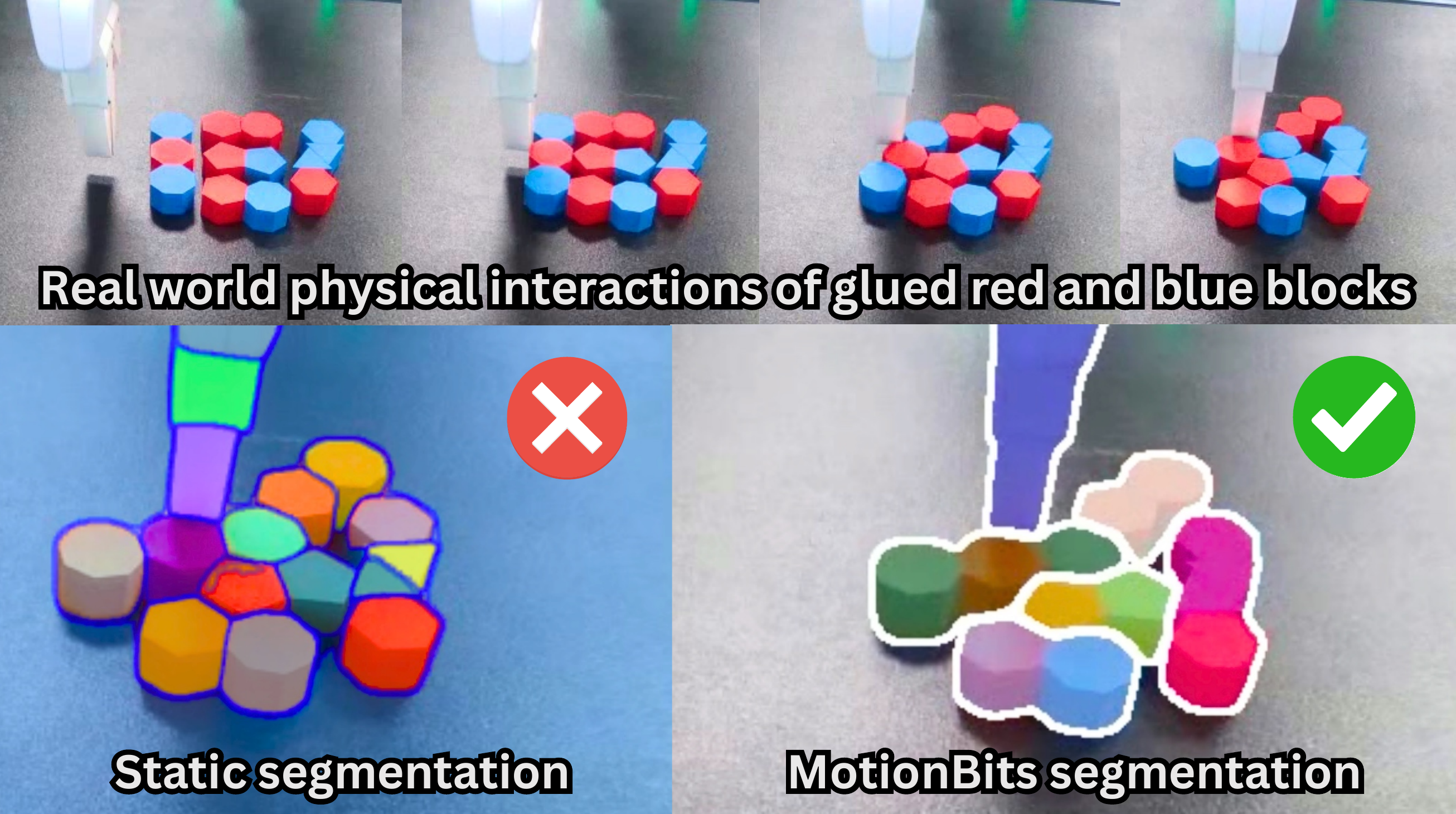}
    \caption{The top row shows a robot physically interacting with a variety of complex composite objects, each constructed from colored blocks that have been glued together. The bottom row highlights the output of a standard semantic segmentation model, which incorrectly over-segments the objects. In contrast, the proposed MotionBits segmentation correctly groups composite objects. Example real-world robotic applications requiring accurate MotionBits segmentation are shown in \cref{fig:application_study_visualizations}.}
    \label{fig:mainfig}
\end{figure}

\section{Introduction}
\label{sec:intro}
Object segmentation across images and videos is a fundamental task in computer vision~\cite{youtubevosbenchmark, sam2}. Current models excel in segmentation based on human-defined semantics~\cite{youtubevos,maskrcnn,yolo,Kirillov2023SegmentA}, which has proven effective for many high level downstream tasks such as classification, tracking, and visual question answering (VQA). However, as vision systems begin to be deployed in the physical world (\eg robotics, VR/AR, video-language models, etc.), they must also be competent for new real-world use cases, where motion-level understanding is imperative~\cite{bohg2017IP}.

Recent segmentation works, including foundation models~\cite{Kirillov2023SegmentA, sam2} and methods using motion as cues~\cite{seganymo, xie2022segmenting, xie2019foreground}, segment the world semantically based on human-defined classes (\eg \textit{desk}, \textit{keyboard}, \textit{microwave}, etc.) that do not take into account how things physically interact in the real world. For example, segmenting a {\em keyboard }does not give us insight to the fact that its keys can be pressed and moved differently relative to the rest of the object. Therefore, while semantics-based segmentation is still important, it is only part of the equation necessary for scene understanding, embodied reasoning, and dexterous manipulation~\cite{bohg2017IP,riseg,rtriseg}. 

Physical interactions often occur at the level of rigid bodies, the most basic manipulable elements of our world. These elements can only be identified via motion analysis~\cite{ModernRobotics}, which means motion-level perception is essential. This perception of moving rigid bodies can enable embodied systems to execute advanced tasks beyond those of simple pick-and-place or shallow VQA. For example, \cref{fig:mainfig} shows how moving rigid-body segmentation can be used for localization of visually complex objects, and \cref{fig:application_study_visualizations} shows how these masks can be used in real-world downstream manipulation tasks, such as tower stacking. 

Additionally, these segmentation masks can be used as cues to enhance vision-language models (VLMs) for more detailed reasoning in visually grounded VQA~\cite{yang2023setofmark}, an increasingly important skill for embodied systems~\cite{song2025robospatial,intelligence2504pi0, fan2025long,geminirobotics, 3dvla, zhao2025cot, huang2025otter, kim24openvla, driess2023palm, belkhale2024minivla, Qwen2.5-VL, damonlpsg2023videollama, internvl}. Furthermore, with the emerging adoption of smart eyeglasses, motion analysis of physical interactions between rigid bodies captured by egocentric videos can be applied to imitation learning, video game simulation, and heads-up task assistance, to name a few~\cite{liu2025egozerorobotlearningsmart,chen2025vidbot,yoshida2025developing,yoshida2025generating,kareer2025egomimic}. Recently, SAPIEN~\cite{xiang2020sapien}, a rigid-part-based robotics simulator, has become a simulator of choice for embodied AI research, serving as the foundation for many datasets, benchmarks, and frameworks, such as ManiSkill~\cite{mu2021maniskill}. This raises a natural question: \textit{if the leading robotics simulator models the world in terms of moving rigid bodies, why shouldn't embodied vision frameworks adopt the same philosophy for segmentation?}

Therefore, we introduce MotionBit, a novel motion-based segmentation concept that, unlike prior works, is formulated through kinematic spatial twist equivalence to capture real-world physical interactions. Tracking these segmentation masks gives insight into how every rigid body in a scene dynamically interacts with one another. In summary, we make the following three main contributions:
\begin{enumerate}
    \item MotionBit: a new concept defining the smallest unit in motion-based segmentation, where each rigid part exhibiting distinct rigid-body motion is assigned a unique mask, independent of semantics. The mathematical definition of a MotionBit is detailed in \cref{sec:motionbits}.
    \item MoRiBo: a new hand-labeled benchmark for evaluating moving rigid-body segmentation across robotic manipulation and human-in-the-wild videos, which is detailed in \cref{sec:benchmark}. Videos for each track are sourced from BridgeData V2~\cite{bridgedatav2} and SA-V~\cite{sam2}, respectively.
    \item A learning-free graph-based method for MotionBits segmentation in dynamic and diverse RGB videos. We evaluate our method's performance on the proposed MoRiBo benchmark via quantitative and qualitative comparisons to state-of-the-art motion-based segmentation models and video-language models. Furthermore, we showcase applications of our MotionBits segmentation in downstream embodied tasks.
\end{enumerate}

\section{Related Works}
\label{sec:related} 
Existing semantic and motion-based segmentation methods learn visual cues but fail to capture the underlying rigid-body dynamics and structures needed for complex manipulation tasks. In this section, we review these approaches, alongside embodied vision systems and relevant datasets, to highlight their limitations in providing low-level physical understanding for intelligent interactions.

\subsection{Semantic and Motion-Based Segmentation}
Segmentation models are often trained on large-scale, semantically-defined datasets \cite{imagenet, pascalvoc, mscoco, cityscapes, guler2018densepose, Kirillov2023SegmentA, sam2, youtubevos, youtubevosbenchmark}. Early approaches focused on classification, detection, and instance segmentation~\cite{fcn, deeplab, maskrcnn, yolo, xianglearningrgbd}, while recent works such as \textit{Segment Anything}~\cite{Kirillov2023SegmentA} and \textit{SAM~2}~\cite{sam2} leverage ``data engines'' to train generalizable and promptable foundation segmentation models. Despite these advancements, such methods rely entirely on human-defined semantics, which omits motion-level awareness critical for reasoning about physical interactions between objects.

In contrast, motion-based approaches utilize temporal and geometric consistency for segmentation. Classical approaches analytically infer structure through association-based motion analysis~\cite{wang1994representing, darrell1991robust, brox2010object, keuper2015motion} or rigid-body transformations~\cite{adiv1985, peleg1990}, while more recent learning-based methods estimate rigid motion cues from RGB-D images and point clouds~\cite{raft3d, motionseg,chao2025segmentationmotionestimationarticulated, zhong2023multi, huang2021multibodysync}. However, these approaches lack kinematic formulations, depend on depth data, or rely on articulation assumptions, ultimately failing to generalize to real-world videos where embodied reasoning occurs. Other works~\cite{xie2022segmenting, xie2019foreground, ranjan2019CC, muthu2020motionseg, seganymo, yang2021self, dave2019towards} incorporate optical flow or alternative motion primitives to guide segmentation but still leak semantics. This occurs either implicitly through a discrete two-slot attention mechanism in the case of Yang \etal~\cite{yang2021self}, or explicitly, through appearance-based training data in the case of Dave \etal~\cite{dave2019towards}. Thus, these motion-based methods overlook rigid-body motion as a fundamental primitive for segmentation in the real world, which limits their capacity to represent physical interactions in dynamic and diverse environments.

\subsection{Vision for Embodied Systems}
Vision is essential for embodied systems and takes many forms, including interactive perception, large multimodal models (LMMs), vision-language-action (VLA) models, and specialized inference techniques. Interactive perception methods guide robot actions to optimize sensory input, and while active perception offers richer visual feedback than passive perception~\cite{bohg2017IP}, these approaches~\cite{riseg, rtriseg, lu2023self} are often tailored to specific physical embodiments, limiting generalization. Demonstration-based datasets have been used to train or fine-tune LMMs and VLA models for scene perception, reasoning, and action planning~\cite{intelligence2504pi0, fan2025long,geminirobotics, 3dvla, zhao2025cot, huang2025otter, kim24openvla, driess2023palm, belkhale2024minivla}. Zhen \etal~\cite{3dvla} and Zhao \etal~\cite{zhao2025cot} propose vision modules which ``imagine'' future image states via diffusion for task planning. Yet, such ``reasoning'' is restricted by the granularity of image predictions, lacks awareness of physical object interactions, and is limited to simple pick-and-place tasks without complex manipulation or spatial requirements. Recently, new inference and training techniques have emerged to enhance perceptual reasoning in LMMs~\cite{xsam, song2025robospatial}. Yang \etal~\cite{yang2023setofmark} show that overlaying label IDs and segmentation masks on images strengthens visual grounding and perceptual reasoning. However, despite recent progress, prior works similarly observe that existing vision techniques within LMMs and VLA models lack motion-level understanding of physical interactions~\cite{siam2025pixfoundation} and have yet to enable dexterous manipulation in complex environments. 

\subsection{Embodied Perception Datasets}
Datasets and benchmarks have driven progress in computer vision, and recently, embodied AI. Modern vision systems emerged from large hand-labeled datasets~\cite{imagenet,pascalvoc,mscoco,cityscapes,guler2018densepose}, which required immense human annotation efforts and enabled models to learn latent feature representations. Similarly, large-scale video datasets advanced video segmentation~\cite{vsb100, davisdataset, youtubevos, youtubevosbenchmark, sam2, segtrackv2}. However, each of these efforts use human-defined semantics and do not explicitly model object-level motion, presenting learning hurdles for robots that must understand how objects physically interact with one another. Shao \etal~\cite{motionseg} introduced a rigid-body segmentation dataset but consisted only of synthetic data, two frames per scene, and assumed non-articulated rigid objects falling onto a tabletop, limiting real-world applications where objects have many rigid parts and are physically interacted with throughout an observation. Recent efforts to assemble large-scale embodied datasets~\cite{openxembodiment, bridgedatav2,song2025robospatial, dong2025digital,mu2025robotwin} have shown promise in robotic perception applications and often fall under two categories: demonstration data via teleoperation and simulation-based trajectory generators. Yet, many such datasets focus manipulation on only one subject per episode and do not offer a general framework for perceiving and understanding the world through interactions. 

\begin{figure}[t]
    \centering
    \includegraphics[width=0.85\linewidth]{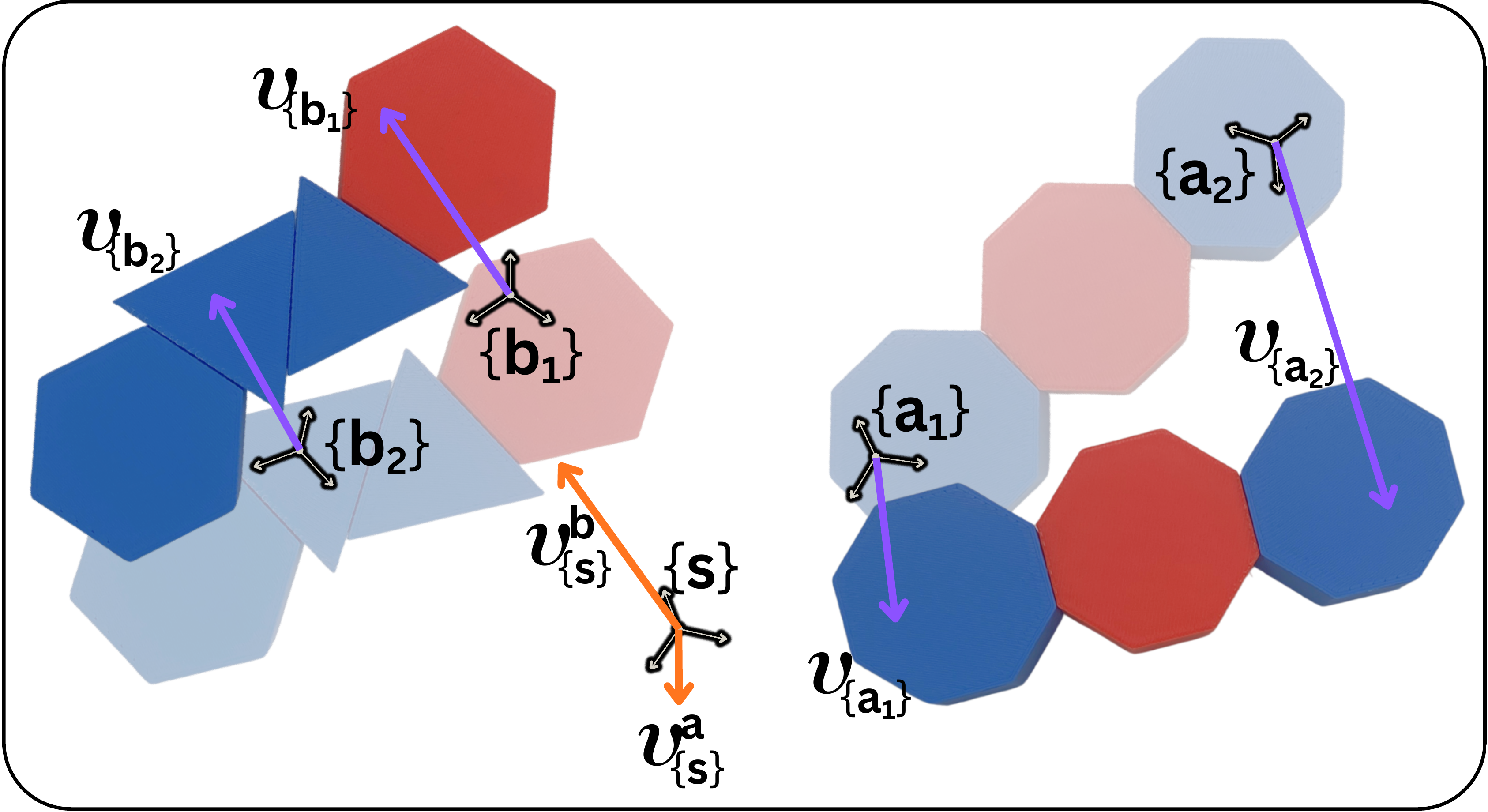}
    \caption{Illustration of spatial twist for two rigid objects. Transparent objects represent initial positions, and solid objects represent positions after motion. The translations of body frames are represented by purple linear velocity vectors $\upsilon_{\{x\}}$. Although $\upsilon_{\{a_1\}} \neq \upsilon_{\{a_2\}}$ and $\upsilon_{\{b_1\}} \neq \upsilon_{\{b_2\}}$, transforming them to the fixed world frame $\{s\}$ yields identical linear velocities $\upsilon_{\{s\}}^x$ for frames on the same rigid object $x$ but distinct linear velocities for different rigid objects. Intuitively, the instantaneous motion observed at the world frame will appear identical for body frames on the same rigid object, regardless of their local motions, but will differ across rigid bodies.}
    \label{fig:motionbits_def}
\end{figure}

\section{The MotionBit Definition}
\label{sec:motionbits}
A MotionBit is a new concept grounded in rigid-body kinematics that represents the smallest unit in motion-based segmentation for understanding physical interactions, where each rigid part exhibiting its own rigid-body motion is assigned a unique segmentation mask, independent of semantics. Accordingly, a MotionBit is formulated to densely map pixels or points to the same label ID if they share the same spatial twist across an observation time window. 

Under rigid-body motion, two body frames, $\{b_1\}$ and $\{b_2\}$, attached to the same rigid body may have different body twists expressed in their local coordinates but will share the same spatial twist when expressed in an arbitrarily chosen fixed world frame $\{s\}$~\cite{ModernRobotics}. Specifically, the instantaneous translation and rotation of an attached body frame $\{b\}$ can be represented as body twist $\mathcal{V}_b = [\omega_b, \upsilon_b]^\intercal \in \mathbb{R}^6$, where angular velocity $\omega_b$ and linear velocity $\upsilon_b$ are represented in the $\{b\}$ frame. 

To derive the corresponding spatial twist, $\mathcal{V}_s = [\omega_s, \upsilon_s]^\intercal \in \mathbb{R}^6$, we use the adjoint representation of the transformation matrix $T_{sb} \in SE(3)$, which defines the configuration of the body frame $\{b\}$ relative to the fixed world frame $\{s\}$. Let $T_{sb}$ be composed of a rotation matrix $R_{sb} \in SO(3)$ and a translation vector $p_{sb} \in \mathbb{R}^3$. The $6 \times 6$ adjoint matrix $[\mathrm{Ad}_{T_{sb}}]$ is thus defined as
\begin{equation}
\label{eq:adjointMatrix}
[\mathrm{Ad}_{T_{sb}}] =
\begin{bmatrix}
R_{sb} & 0_{3 \times 3} \\
[p_{sb}]R_{sb} & R_{sb}
\end{bmatrix},
\end{equation}
where $[p_{sb}]$ is the $3 \times 3$ skew-symmetric matrix representation of $p_{sb}$. We then map the body twist to the spatial frame directly via
\begin{equation}
\label{eq:twistDerivation}
\mathcal{V}_s = [\mathrm{Ad}_{T_{sb}}] \mathcal{V}_b.
\end{equation}
\indent Afterwards, the spatial twist $\mathcal{V}_s$ of each body frame is re-expressed about the origin of the fixed world frame $\{s\}$, instead of the body frame $\{b\}$, to enable kinematic comparison across body motions. When expressing a body's spatial twist about $\{s\}$, the angular velocity $\omega_s$ remains unchanged. However, the linear velocity must be shifted by $-\omega_s \times p_{sb}$ to account for the offset between the origins of $\{s\}$ and $\{b\}$, where $p_{sb}$ is the translation vector from $\{s\}$ to $\{b\}$ expressed in the $\{s\}$ frame. For brevity, we hereafter denote this simply as the spatial twist $\mathcal{V}_s$. In this representation, spatial twists of body frames of the same rigid body will be identical. \cref{fig:motionbits_def} visualizes the physical interpretation of spatial twist.

Extending beyond an instantaneous motion description, a MotionBit enforces temporal consistency of spatial twists. Specifically, two pixels or points, $x_i$ and $x_j$, belong to the same MotionBit if their spatial twists remain kinematically equivalent throughout an observation time window $\mathcal{T} = [t_{start}, t_{end}]$, defined by
\begin{equation}
\label{eq:temporal_twist}
\|\Delta\mathcal{V}_s(x_i,x_j,t)\|_2 = \|\mathcal{V}_s(x_i, t) - \mathcal{V}_s(x_j, t)\|_2 = 0, \quad \forall t \in \mathcal{T},
\end{equation}
where $\mathcal{V}_s(x, t)$ denotes the spatial twist of $x$ at time $t$ and $\Delta\mathcal{V}_s(x_i,x_j,t)$ denotes the difference between two spatial twists at time $t$. Since a MotionBit must exhibit motion during the observation window, we define this constraint as
\begin{equation}
\label{eq:twist_motion_constraint}
\exists\, t \in \mathcal{T} \;\; \text{ s.t. } \;\; \|\mathcal{V}_s(x,t)\|_2 \neq 0 \quad \forall x \in M ,
\end{equation}
where points $x$ in MotionBit $M$ also pairwise satisfy \cref{eq:temporal_twist}. Thus, each MotionBit is composed of a set of points that share an identical non-zero spatial twist trajectory, representing a rigid part in the scene that has exhibited some motion. This is formally defined by \cref{def:MotionBit}.
\begin{samepage}
\begin{definition}[MotionBit]
\label{def:MotionBit}
Let $X$ be the set of all observed points and $\mathcal{V}_s(x^*,t)$ denote the spatial twist of a reference point $x^* \in X$ at time $t$, such that $\|\mathcal{V}_s(x^*,t)\|_2 \neq 0$ for some $t$ in the observation time window $T$. We define the set $M \subseteq X$ as
\[
M = \big\{x \in X \;\;\big|\;\; \|\Delta\mathcal{V}_s(x^*, x, t)\|_2 = 0,\;\; \forall t \in T \big\},
\]
which satisfies the constraints in \cref{eq:temporal_twist,eq:twist_motion_constraint} and constitutes one \textbf{MotionBit}.
\end{definition}
\end{samepage}

\begin{table}[b]
  \centering
  \renewcommand{\arraystretch}{0.9}
    \setlength{\aboverulesep}{2pt}
    \setlength{\belowrulesep}{2pt}
    \setlength{\extrarowheight}{2pt}
    \caption{Statistics of the MoRiBo benchmark for evaluating moving rigid-body segmentation. Each video has resolution of 480$\times$640 or 640$\times$480, and hand-annotated segmentation masks are provided for the last frame of each video.}
    \label{tab:benchmark_table}
    \scalebox{0.9}{%
  \begin{tabular}{@{}lccccc@{}}
    \toprule
    Track & \#Scenes & Frames/Scene & Masks/Scene & Source Dataset & Actions and Motions \\
    \midrule
    Robotic Manipulation & 270 & $15.9 \pm 0.7$ & $5.2 \pm 1.5$ & BridgeData V2~\cite{bridgedatav2} & Push, Grasp, Pick-\&-Place \\
    Human-in-the-Wild & 79 & $11.0 \pm 0.0$ & $10.3 \pm 8.6$ & SA-V~\cite{sam2} & Human-Object Interactions \\
    \bottomrule
  \end{tabular} }
\end{table}

\begin{figure}[t]
    \centering
    \includegraphics[width=0.85\linewidth]{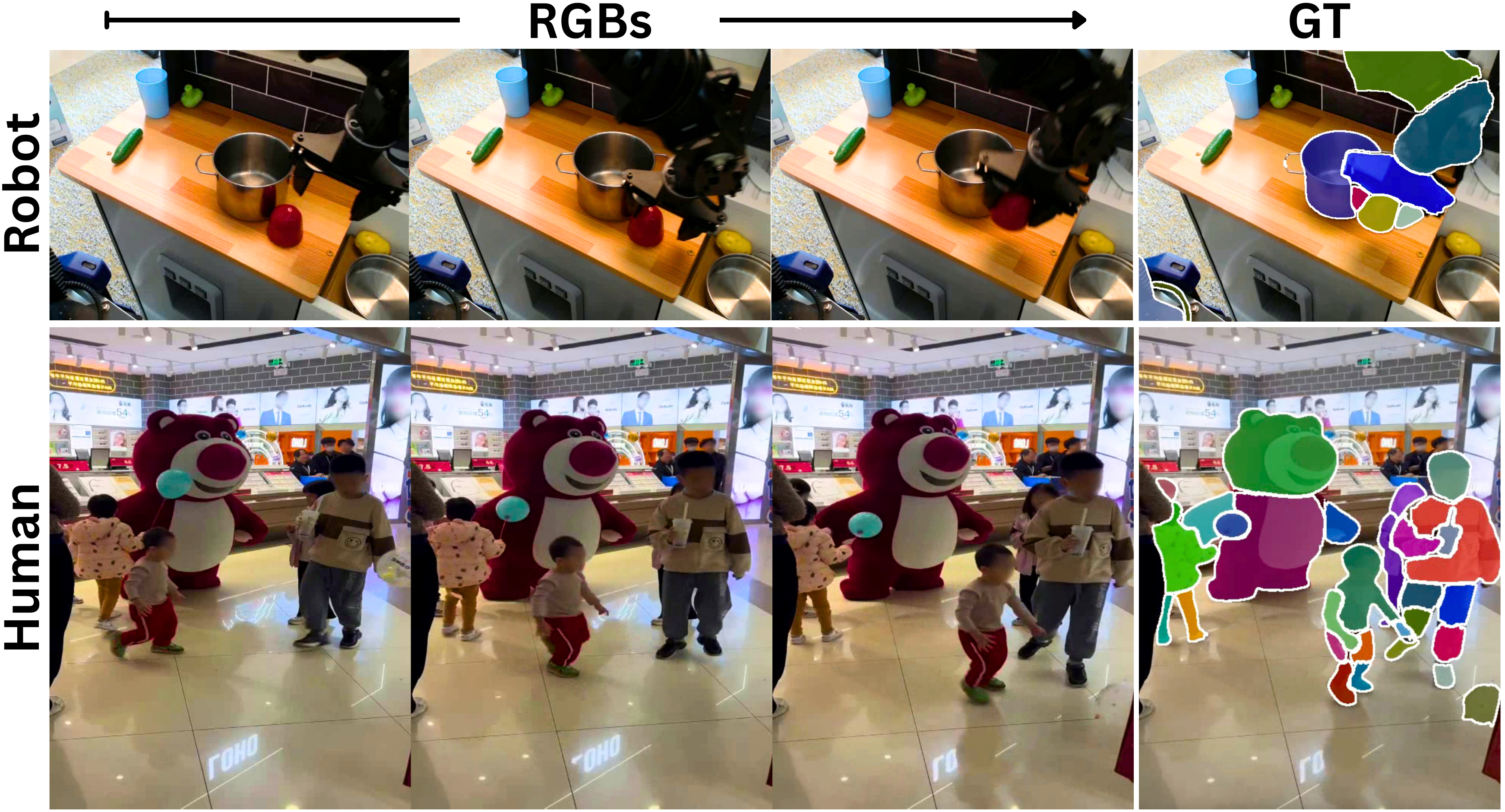}
    \caption{Examples from the new MoRiBo benchmark across both tracks, robotic manipulation and human-in-the-wild. Hand-labeled final-frame segmentation masks of moving rigid bodies are provided for every video.}
    \label{fig:benchmark_examples}
\end{figure}

\section{The MoRiBo Benchmark}
\label{sec:benchmark}
The new MoRiBo benchmark is the first evaluation framework for real-world moving rigid-body segmentation in RGB videos, providing baselines for low-level physical interaction understanding. It consists of 270 robotic manipulation videos from BridgeData V2~\cite{bridgedatav2} and 79 human-in-the-wild videos from SA-V~\cite{sam2}, organized into two tracks to support moving rigid-body segmentation across domains. Videos from BridgeData V2 are captured from a static camera and depict a teleoperated robot arm performing pushing, grasping, and pick-and-place tasks in a lab setting, while videos from SA-V are recorded with minimal camera motion and depict in-the-wild human-object interactions. This setup ensures that motion cues primarily correspond to dynamic rigid bodies, rather than camera ego-motion. \cref{tab:benchmark_table} summarizes the MoRiBo benchmark statistics and \cref{fig:benchmark_examples} provides qualitative examples of moving rigid-body scene annotations.

Ground-truth annotations were manually generated using SAM2-assisted labeling, where each rigid part exhibiting independent motion was assigned a unique segmentation mask. Specifically, each entry underwent an intensive process of manual refinements to initial human-in-the-loop SAM2 predictions, which ensured that boundaries are kinematically accurate according to \cref{def:MotionBit}. The benchmark provides only ground-truth masks for the last frame of each video and encourages researchers to apply their own rigid-body motion analysis techniques or propagate masks if needed. Since only final-frame masks are provided, a variety of approaches, from learning-free motion clustering frameworks to transformer-based video models, can be fairly compared for their ability to identify rigid parts across an observation time window. Thus, the proposed MoRiBo benchmark is designed purely for evaluation, and segmentation performance is measured using Overlap and Boundary metrics, further detailed in \cref{sec:experiments}. Both the benchmark and evaluation code will be publicly released to support further research in motion-centric segmentation and embodied vision.

\begin{figure}[t]
\begin{algorithm}[H]
\resizebox{0.85\columnwidth}{!}{%
    \begin{minipage}{1.0\columnwidth}
\setstretch{1.0} 
\caption{Graph-Based MotionBits Segmentation}
\label{alg:motionbits_loop}
\textbf{Input:} RGB video stream, $V$ \\
\textbf{Output:} MotionBits segmentation masks, $\mathcal{L}$
\begin{algorithmic}[1]

\State Initialize: $t \gets 0$
\State Initialize empty sets: $\mathcal{I} \gets \emptyset$, $\mathcal{G} \gets \emptyset$, $\mathcal{L} \gets \emptyset$
\While{$V$ has next frame $I_t$}
    \If{$\mathcal{I} \neq \emptyset$}
        \State $P_t, Q_t \gets \textsc{SampleNodes}(I_t, \mathcal{I})$ \Comment{\cref{sec:method_graph_construct}}
        \State $\mathcal{V}_s \gets \textsc{EstimateLocalTwists}(P_t, Q_t)$ \Comment{\cref{sec:method_graph_construct}}
        \State $G_t \gets \textsc{BuildTwistSimGraph}(\mathcal{V}_s, I_t, \mathcal{I}, \mathcal{G}, \mathcal{L})$ \Comment{\cref{sec:method_graph_construct}}
        \State $B_t \gets \textsc{SoftLabelPropagation}(G_t)$ \Comment{\cref{sec:method_soft_hard_seg}}
        \State $L_t \gets \textsc{HardMarkovClustering}(B_t)$ \Comment{\cref{sec:method_soft_hard_seg}}
        \State $\mathcal{G} \gets \mathcal{G} \cup \{G_t\}$; \quad $\mathcal{L} \gets \mathcal{L} \cup \{L_t\}$
        \State \textbf{yield} $L_t$
    \EndIf
    \State$\mathcal{I} \gets \mathcal{I} \cup \{I_t\}$
    \State $t \gets t + 1$
\EndWhile
\State \textbf{return} $\mathcal{L}$
\end{algorithmic}
\end{minipage}
  }
\end{algorithm}
\end{figure}

\section{Graph-Based MotionBits Segmentation}
\label{sec:method}
\begin{figure*}[b]
    \centering
    \includegraphics[width=1.0\linewidth]{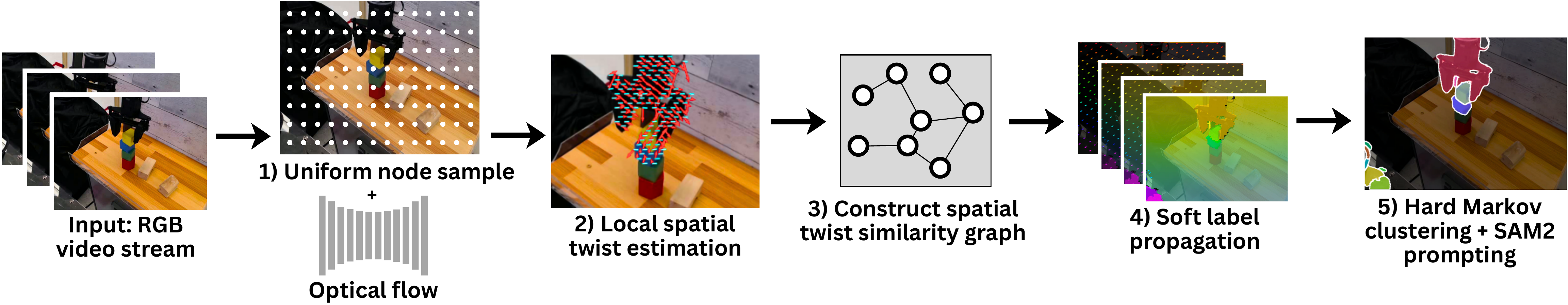}
    \caption{Our learning-free graph-based method for online MotionBits segmentation.}
    \label{fig:method_visualized}
\end{figure*}

Along with the MotionBit concept and the MoRiBo benchmark, we propose a learning-free graph-based method for MotionBits segmentation, outlined by \cref{alg:motionbits_loop} and illustrated by \cref{fig:method_visualized}.

The input to our method is a video stream, $V = \{I_t\}_{t=0}^T$, where $t \in \mathbb{Z}_{\geq0}$ is the discrete time step, $T\in \mathbb{Z}_+$ is an arbitrary number of total time steps, and $I_t \in [0,255]^{H \times W \times 3}$ is the RGB image at $t$. For each frame $I_{t}$ at $t\geq 1$, our framework yields a MotionBits segmentation mask $L_t \in \mathbb{Z}_{\geq 0}^{H \times W}$ in an online manner using only images $\{I_\tau\}_{\tau=0}^t$, where the object ID given by $L_t^{(x,y)}$ corresponds to the pixel $(x,y)$ in image $I_t$. If $L_t^{(x,y)}=0$, pixel $(x,y)$ belongs to the background and is not part of a rigid body that exhibited motion up to time step $t$. Values $L_t^{(x,y)} > 0$ indicate pixel $(x,y)$ belongs to a MotionBit with unique ID $L_t^{(x,y)}$.

\subsection{MotionBits Graph Construction}
\label{sec:method_graph_construct}
Given an RGB video stream $V = \{ I_t \}_{t=0}^{T}$, our framework constructs a spatial twist similarity graph for each new image frame to model local rigid-body motions. The graph $G_t = (\mathcal{M}_t, \mathcal{E}_t, \mathcal{W}_t)$ is constructed in lines 5-7 of \cref{alg:motionbits_loop} and serves as the structural foundation for both soft and hard MotionBits segmentation. Although the MotionBit is defined in SE(3), we choose to implement our moving rigid body segmentation method in SE(2) to prioritize compatibility with RGB videos, which is the most prevalent visual perception channel across robotics and computer vision. A sensitivity analysis is provided at the end of this section to justify the robustness of this decision.

At each time step $t \geq 1$, we obtain forward and backward optical flow fields, $F_{t-1 \rightarrow t}, \; F_{t \rightarrow t-1} \in \mathbb{R}^{H \times W \times 2}$, between frames $I_{t-1}$ and $I_t$ via an off-the-shelf optical flow model~\cite{dong2024memflow}. In $\textsc{SampleNodes}(\cdot)$, a uniformly sampled $\sqrt{n} \times \sqrt{n}$ grid of $n$ total points across the image plane produces matched point sets
\begin{equation}
P_t = \{ p_{(i,j)} \}, \quad
Q_t = \{ q_{(i,j)} = p_{(i,j)} + F_{t \rightarrow t-1}(p_{(i,j)}) \},
\end{equation}
where points in $P_t$ lie in the current frame $I_t$ and define graph nodes $m_{(i,j)}\in \mathcal{M}_t$, and points in $Q_t$ correspond to the points in $P_t$ tracked backward to $I_{t-1}$. By using $P_t$ as nodes and reverse motions for graph construction, we ensure all regions are covered in the image space of $I_t$.

For each grid point $(i,j)$, its $k$-nearest neighbors is defined as the set $\mathcal{N}_{(i,j)}$. $\textsc{EstimateLocalTwists}(\cdot)$ then estimates the local spatial twist by solving for the rigid-body motion between $p_{(i,j)}$ and its corresponding backwards projected point $q_{(i,j)}$. By using $\mathcal{N}_{(i,j)}^* = \mathcal{N}_{(i,j)} \cup \{(i,j)\}$, the rigid-body motion that transforms $P_{\mathcal{N}_{(i,j)}^*}$ to $Q_{\mathcal{N}_{(i,j)}^*}$ is defined by
\begin{equation}
Q_{\mathcal{N}_{(i,j)}^*} = R P_{\mathcal{N}_{(i,j)}^*} + \boldsymbol{t},
\label{eq:rp_t}
\end{equation}
where $P_{\mathcal{N}_{(i,j)}^*}$ is the set of points $\mathcal{N}_{(i,j)}^* \subset P_t$, $Q_{\mathcal{N}_{(i,j)}^*}$ is the corresponding set of points in $Q_t$, $R$ is a rotation matrix, and $\boldsymbol{t}$ is the translation vector. Using modified RANSAC with weighted Kabsch estimation, we compute the local rotation angle $\omega$ and represent the resulting body twist as $\mathcal{V}_{b}(i,j) = [\omega, v_x, v_y]$.

The origin of the image coordinate system is selected to be the fixed world frame $\{s\}$, and transforming each body twist into this shared frame yields the spatial twist $\mathcal{V}_{s}(i,j)$, which characterizes each node during graph construction. All features are computed with respect to the selected fixed world frame for consistency and simplified computation.

Then, in line 7 of \cref{alg:motionbits_loop}, $\textsc{BuildTwistSimGraph}(\cdot)$ constructs a similarity graph $G_t$, where each node $m_{(i,j)} \in \mathcal{M}_t$ corresponds to a grid location $(i,j) \in  P_t$ with spatial twist $\mathcal{V}_s(i,j)$. Similarity edges $\mathcal{E}_t$ and corresponding weights $\mathcal{W}_t$ are defined using a Gaussian kernel over the Mahalanobis distance~\cite{Mahalanobis}, denoted by $\mathrm{Mdist}(\cdot)$. For each node $(i,j)$, we compute a covariance matrix $\Sigma_{\mathcal{N}_{(i,j)}^*}$ over its neighborhood spatial twists $\{\mathcal{V}_s(a,b)\}_{(a,b)\in \mathcal{N}_{(i,j)}^*}$ and define pairwise similarities between $(i,j)$ and all $(a,b) \in \mathcal{N}_{(i,j)}$ via
\begin{gather}
\mathrm{Mdist}((i,j),(a,b)) = \sqrt{ (\Delta \mathcal{V}_s)^\top \Sigma_{\mathcal{N}_{(i,j)}^*}^{-1} (\Delta \mathcal{V}_s) },\\
\mathcal{W}_t^{(i,j),(a,b)} = \exp\!\left( -\tfrac{1}{2} \mathrm{Mdist}((i,j),(a,b))^2 \right).
\end{gather}

To maintain temporal consistency, prior segmentation masks $L_{\tau < t} \in \mathcal{L}$ and graphs $G_{\tau < t} \in \mathcal{G}$ are integrated to modify graph $G_t$ by adding and subtracting edges from $\mathcal{E}_t$. Each previous mask $L_\tau$ is projected forward to time $t$ using the forward flows $F_{\tau \rightarrow \tau+1}...F_{t-1 \rightarrow t}$. First, edges are added if two nodes belong to the same projected MotionBit for all $\tau<t$, with edge weight equal to the greatest $\mathcal{W}_\tau ^{(i,j),(a,b)}$ from projected graphs $\mathcal{G}_{\tau < t}$. Edges are then removed if two nodes are separated across any $\tau\leq t$. The new graph $G_t$ therefore encodes spatial twist similarity in a manner temporally consistent with the observed motions of the scene. It should be noted that our method assumes a primarily static camera, where motion at the scale of optical flow noise can be reasonably accounted for via the locally adaptive Mahalanobis kernel. We leave full SE(3) camera compensation to future work.

\subsubsection{SE(2) Model Justification.}
\label{subsec:se(2)just}
We quantitatively validate the robustness of the chosen SE(2) motion model in \cref{eq:rp_t} through a sensitivity analysis designed to measure its invariance to spatial perturbations of an observed object. We define the sensitivity matrix $\Psi$ as the concatenation of the spatial partial derivatives of the image interaction matrix $\mathbb{L}$~\cite{chaumette2006visual}, where each derivative is projected by an unconstrained SE(3) spatial twist, $\mathcal{V}_{SE(3)}$. By using $\Psi$, we can predict the projected spatial twist difference $\delta\mathcal{V}_{SE(2)}$ between two points on the same moving rigid body separated by a 3D displacement $\delta\mathbf{p}$. This quantifies the extent to which the SE(2) model deviates from the true SE(3) motion as the distance between tracked points varies.
\begin{equation}
\delta \mathcal{V}_{SE(2)} \approx \left[ \frac{\partial \mathbb{L}}{\partial X}\mathcal{V}_{SE(3)} \;\; \frac{\partial \mathbb{L}}{\partial Y}\mathcal{V}_{SE(3)} \;\; \frac{\partial \mathbb{L}}{\partial Z}\mathcal{V}_{SE(3)} \right] \delta \mathbf{p} = \Psi \delta \mathbf{p}.
\end{equation}

To quantify SE(2) deviation with respect to corresponding SE(3) motion, we perform a Monte Carlo simulation consisting of 100,000 trials. For each trial, we generate a rigid object under unconstrained 3D motion by randomly sampling SE(3) spatial twists and uniformly sampling the direction of the 3D displacement vector $\delta \mathbf{p}$ between two tracked points. We then simulate the full perspective projection of these points to measure the empirical kinematic variation.

First, for a typical robotic workspace, we consider a rigid object positioned within $X,Y\in[\pm2\mathrm{m}]$ and $Z=1.5\mathrm{m}$ in the camera frame. Because our method densely samples points in the image plane, we evaluate the SE(2) model with a point separation of $\|\delta\mathbf{p}\|=2\mathrm{cm}$. Given these SE(3) parameters, we obtain a relative difference of $\delta\mathcal{V}_{SE(2)}^{rel}=0.897\% \pm 0.525\%$ between the observed SE(2) spatial twists of the two points. Furthermore, for in-the-wild scenarios, we consider a rigid object at $X,Y\in[\pm6\mathrm{m}]$, $Z=6\mathrm{m}$, with a larger point separation of $\|\delta\mathbf{p}\|=8\mathrm{cm}$. These parameters yield a relative difference of $\delta\mathcal{V}_{SE(2)}^{rel}=0.752\% \pm 0.455\%$. 

Since an average kinematic error below 1\% is heavily eclipsed by standard optical flow noise and camera jitter, the loss in theoretical precision is negligible. Therefore, these empirical bounds justify reducing the 6-DoF SE(3) problem to an SE(2) motion model, enabling applications across diverse RGB videos.

\begin{figure}[b]
    \centering
    \includegraphics[width=1.0\linewidth]{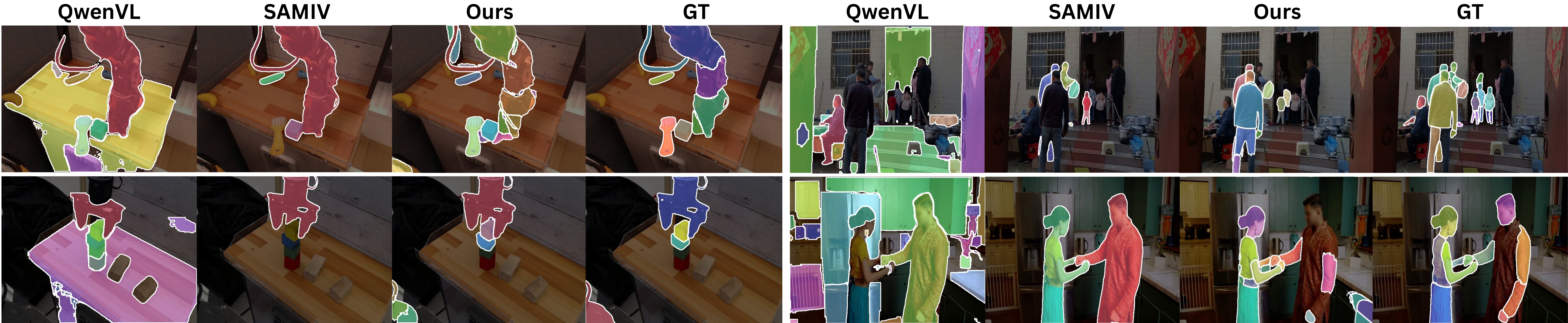}
    \caption{Qualitative comparison of moving rigid-body segmentation on the two-track MoRiBo benchmark between Qwen2.5-VL (QwenVL)~\cite{Qwen2.5-VL}, Segment Any Motion in Videos (SAMIV)~\cite{seganymo} and our method.}
    \label{fig:qualitative_benchmark_results}
\end{figure}

\subsection{Soft and Hard MotionBits Segmentation}
\label{sec:method_soft_hard_seg}
In lines 8-9 of \cref{alg:motionbits_loop}, we perform MotionBits segmentation on the defined similarity graph $G_t = (\mathcal{M}_t, \mathcal{E}_t, \mathcal{W}_t)$ by introducing controlled nonlinearities through a two-stage process. Drawing inspiration from Zhou \etal~\cite{label_propagation}, soft label propagation first diffuses local affinities into a smooth global embedding. After which, hard Markov clustering discretizes this representation into coherent MotionBit segments. For the rest of this section, subscript $t$, denoting the current time step, will be dropped for readability.

For graph $G = (\mathcal{M}, \mathcal{E}, \mathcal{W})$ with $n = |\mathcal{M}|$ nodes and nonnegative affinity matrix $\mathcal{W}\in[0,1]^{n\times n}$, $\textsc{SoftLabelPropagation}(\cdot)$ first selects a small uniformly spaced set of seed nodes $S\subset \mathcal{M}$ and assigns each seed a unique class label among $\{1,\dots,C\}$. The seed labeling is encoded by one-hot matrix $Y\in\{0,1\}^{n\times C}$, where%
\begin{equation}
\label{eq:paint_color_encoding}
Y_{i,c}=
\begin{cases}
1, \; \text{if node } i \in S \text{ and has label } c\\
0, \; \text{otherwise.}
\end{cases}
\end{equation}

Using the initial seeding matrix $Y$ and the affinity matrix $\mathcal{W}$, we diffuse local spatial twist similarities into a smooth, globally consistent embedding. Intuitively, the diffusion behaves as if we pour a unique color of paint on each seed node and allow the colors to flow and blend naturally according to the graph structure, resulting in softly color-coded MotionBits that capture coherent motion regions.

Formally, let $D=\mathrm{diag}(\sum_j \mathcal{W}_{ij})$ be the degree matrix and $A = D^{-1}\mathcal{W}$ be the row-stochastic transition matrix of a random walk on $G$. Then, starting from $B^{(0)}=Y$, we begin diffusing the ``paint'' initialized at each seed node, encoded through $Y$, by iterating
\begin{equation}
B^{(r+1)} = AB^{(r)}.
\end{equation}
Periodically, we ``pour more paint'' onto the seed nodes by re-enforcing initial seed constraints every $r^*$ iterations via
\begin{equation}
B^{(r)} \leftarrow B^{(r)} +  Y.
\end{equation}
After $R$ iterations, we obtain the soft label embedding $B^{(R)}\in \mathbb{R}_{\geq 0}^{n\times C}$.

$\textsc{HardMarkovClustering}(\cdot)$ then discretizes this representation into MotionBits segmentation. We begin by row-wise $\ell_2$-normalization of the soft label embedding, denoted as $\hat B$. Then, we form a symmetric similarity matrix representation of our graph $G$ via inner products
\begin{equation}
X = \hat{B}\hat{B}^{\top} \in \mathbb{R}^{n\times n}.
\end{equation}
Unit self-loops and column-normalization are applied to $X$ to obtain a column stochastic matrix $\hat X$. The Markov clustering algorithm on matrix $\hat X$ is then used to cluster nodes of graph $G$. Intuitively, Markov clustering simulates random walks within a graph with the idea that, over time, these walks will cross over nodes with stronger connections more often than those with weaker connections. Markov clustering is well documented, and the technical details will be omitted in this paper. Once node clusters are extracted via Markov clustering, Segment Anything Model 2~\cite{sam2} is prompted for refined boundaries, and the MotionBits segmentation mask $L_t$ is yielded.

\begin{table}[t]
\centering
\caption{Quantitative results on the MoRiBo benchmark (\%). Our method surpasses alternatives across both the Robotic Manipulation and Human-in-the-Wild tracks.}
\label{tab:quantitative_benchmark_results}
\setlength{\tabcolsep}{10pt}
\renewcommand{\arraystretch}{0.95}
\setlength{\aboverulesep}{0pt}
\setlength{\belowrulesep}{0pt}
\setlength{\extrarowheight}{0pt}
\scalebox{0.9}{%
\begin{tabular}
{
c l cccc ccc}
\toprule
\multicolumn{2}{c}{} & \multicolumn{4}{c}{\bf Overlap} & \multicolumn{3}{c}{\bf Boundary} \\
\cmidrule(lr){3-6} \cmidrule(lr){7-9}
\bf  & \bf Method & P & R & $\text{F}_1$ & mIoU & P & R & $\text{F}_1$ \\
\midrule
\multirow{7}{*}{\rotatebox[origin=c]{90}{\shortstack{Robotic\\Manipulation}}}
 & VideoLLaMa~\cite{damonlpsg2023videollama} & 47.2 & 9.2 & 8.6 & 7.4 & 43.8 & 8.4 & 7.7 \\
 & InternVideo~\cite{internvl} & 49.5 & 17.7 & 15.0 & 13.9 & 44.2 & 14.2 & 13.8 \\
 & QwenVL~\cite{Qwen2.5-VL} & 19.1 & 28.7 & 24.6 & 21.5 & 17.7 & 25.1 & 21.9 \\
 & SAMIV~\cite{seganymo} & {\bf 69.5} & 31.0 & 27.0 & 24.3 & {\bf 61.6} & 27.2 & 25.8 \\
 & OCLR Flow~\cite{xie2022segmenting} & 52.5 & 16.9 & 24.4 & 10.2 & 23.0 & 14.3 & 17.8 \\
 & OCLR TTA~\cite{xie2022segmenting} & 60.1 & 19.7 & 28.5 & 12.1 & 32.6 & 11.5 & 16.1 \\
 \cmidrule(lr){2-9}
 & \violetcell{\textbf{Ours}} & \violetcell{45.6} & \violetcell{\textbf{62.7}} & \violetcell{\textbf{59.9}} & \violetcell{\textbf{52.6}} & \violetcell{40.2} & \violetcell{\textbf{53.7}} & \violetcell{\textbf{52.0}} \\

\midrule
\multirow{7}{*}{\rotatebox[origin=c]{90}{\shortstack{Human\\-in-the-Wild}}}
 & VideoLLaMa~\cite{damonlpsg2023videollama} & 48.9 & 12.8 & 9.1 & 8.5 & 44.5 & 10.9 & 8.6 \\
 & InternVideo~\cite{internvl} & 44.3 & 14.8 & 10.7 & 8.4 & 40.0 & 11.6 & 9.8 \\
 & QwenVL~\cite{Qwen2.5-VL} & 18.3 & 25.2 & 16.2 & 13.4 & 16.8 & 23.3 & 16.7 \\
 & SAMIV~\cite{seganymo} & {\bf 51.5} & 18.1 & 13.1 & 11.2 & 42.6 & 14.4 & 11.4 \\
 & OCLR Flow~\cite{xie2022segmenting} & 42.0 & 11.9 & 16.3 & 7.4 & 19.9 & 12.6 & 13.2 \\
 & OCLR TTA~\cite{xie2022segmenting} & 48.8 & 13.6 & 19.8 & 9.9 & 26.0 & 10.0 & 13.2 \\
 \cmidrule(lr){2-9}
 & \violetcell{\textbf{Ours}} & \violetcell{49.2} & \violetcell{\textbf{53.5}} & \violetcell{\textbf{52.4}} & \violetcell{\textbf{46.7}} & \violetcell{\textbf{45.9}} & \violetcell{\textbf{49.7}} & \violetcell{\textbf{47.4}} \\
\bottomrule
\end{tabular}}
\end{table}

\section{Experiments}
\label{sec:experiments}
In this section, we use the MoRiBo benchmark to quantitatively (\cref{tab:quantitative_benchmark_results}) and qualitatively (\cref{fig:qualitative_benchmark_results}) compare our proposed method against state-of-the-art embodied intelligence and motion segmentation baselines in moving rigid-body segmentation. Acknowledging that these methods were not specifically designed for moving rigid-body segmentation, we further provide qualitative demonstrations in downstream applications where current vision systems fail to provide essential cues for embodied reasoning and dexterous manipulation, highlighting the necessity for MotionBits segmentation in complex real-world environments.

\subsection{Experiment Setup}
\label{subsec:experiment_setup}
We compare our method against baselines by evaluating performance on the MoRiBo benchmark, introduced in \cref{sec:benchmark}. Baselines fall under two major categories: (1) video-language models, which include VideoLLaMA3-7B~\cite{damonlpsg2023videollama}, Qwen2.5-VL-32B~\cite{Qwen2.5-VL}, and InternVideo2.5-8B~\cite{internvl}, and (2) motion segmentation methods, which include Segment Any Motion in Videos (SAMIV)~\cite{seganymo}, OCLR Flow (flow only)~\cite{xie2022segmenting}, and OCLR TTA (flow with test-time adaptation)~\cite{xie2022segmenting}.

Because the evaluated VLMs do not natively produce pixel-wise multi-object segmentation masks, we employ YOLOE-L~\cite{wang2025yoloerealtimeseeing}, a SOTA promptable open-vocabulary segmentation model, to translate textual VLM responses to segmentation masks. Each VLM is prompted with:
\begin{samepage}
\begin{quote}
\small
``\textit{What rigid bodies or parts have moved in the video? Describe each in detail.}''
\end{quote}
\end{samepage}
The prompt is accompanied with output format specifications to ensure compatibility with the YOLOE model.

For each video, the final-frame segmentation mask produced by each method is evaluated against the manually annotated ground-truth mask using Hungarian matching. Performance metrics include precision (P), recall (R), and $\text{F}_1$ score computed under both Overlap and Boundary criteria. The Overlap criterion additionally includes mean intersection-over-union (mIoU). All metrics are macro-averaged so that each instance is weighted equally, regardless of its pixel area.

\begin{figure*}[t]
    \centering
    \includegraphics[width=0.85\linewidth]{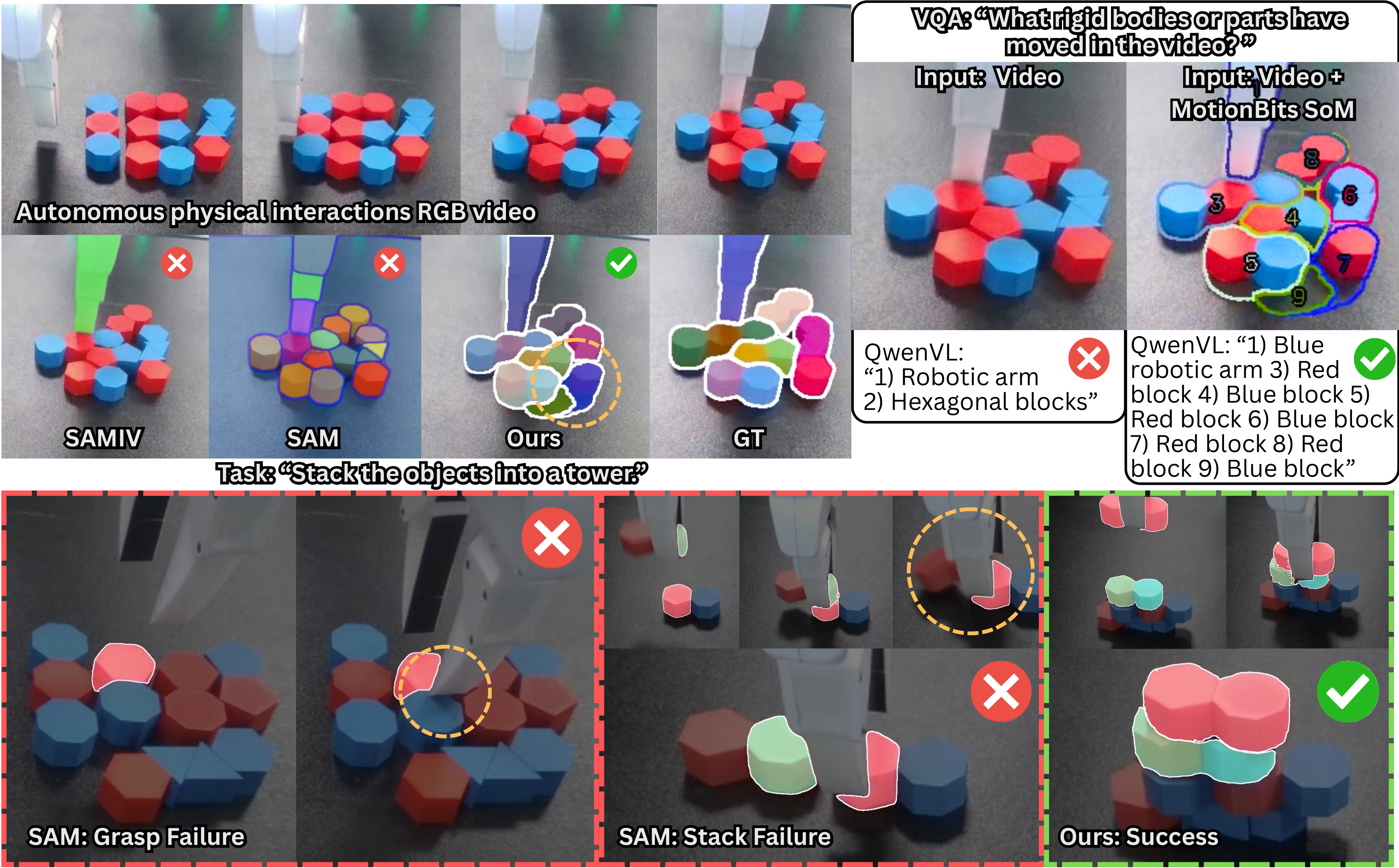}
    \caption{Application examples of MotionBits for downstream tasks in a tabletop scene where a robot autonomously interacts with complex, composite objects. QwenVL~\cite{Qwen2.5-VL} fails to produce an accurate response in the VQA task. However, using our MotionBits segmentation as overlaid marks~\cite{yang2023setofmark} improves VQA response accuracy. Comparing object segmentation among SAMIV~\cite{seganymo}, SAM~\cite{Kirillov2023SegmentA}, and our MotionBits method shows that incorporating rigid-body motions as segmentation cues gives more accurate object-level insight. For the tower stacking task, over-segmentation of the composite objects by SAM~\cite{Kirillov2023SegmentA} generates robot actions that result in grasp and stack failures. In contrast, our method's MotionBits segmentation enables successful grasping and stacking.}
    \label{fig:application_study_visualizations}
\end{figure*}

\subsection{Evaluation Results and Discussion}
In \cref{tab:quantitative_benchmark_results}, we demonstrate our method's superior performance on the MoRiBo benchmark against SOTA video-language models and motion segmentation methods. We consistently outperform baselines in both tracks across all major metrics in Overlap $\text{F}_1$, Overlap mIoU and Boundary $\text{F}_1$. On average, our method outperforms all baselines by 37.3\% mIoU and outperforms the two best performing baselines (Qwen2.5-VL and Segment Any Motion in Videos) by 32.1\% mIoU, across both tracks. Additionally, we qualitatively compare our method's segmentation masks with Qwen2.5-VL and Segment Any Motion in Videos in \cref{fig:qualitative_benchmark_results}.

These experiments show that baseline vision methods lack perception grounded in physical interactions. While their perception modules encode semantic information, they neglect motion-level cues that reveal the underlying structure of object classes, which is especially critical in out-of-distribution scenarios. In the following subsection, we demonstrate why MotionBit perception is necessary for guiding embodied reasoning and manipulation in novel, unseen environments.

\subsection{MotionBits Downstream Applications}
Recently, VLMs have been increasingly employed to fine-tune VLA models, enabling them to function as embodied reasoning modules in robots. While these models provide strong semantic scene understanding, effective real-world manipulation requires precise reasoning about the geometry and dynamics of complex, unseen objects. In \cref{fig:application_study_visualizations}, red and blue blocks are glued together to form composite objects and placed on a tabletop. Then, an RGB video is taken of a robot autonomously interacting with the cluttered scene.

When prompted to identify the rigid objects in the video that were physically interacted with, QwenVL exhibits a fundamental failure in leveraging motion cues for accurate scene reasoning. Additionally, existing segmentation models fail to accurately segment the objects due to heavy bias toward semantic cues. Static segmentation model, SAM, over-segments these objects, and motion-based segmentation model, SAMIV, under-segments. In contrast, our MotionBits segmentation method accurately segments the objects using motion cues from physical interactions. Finally, by overlaying our segmentation masks as visual marks on the RGB video input, we provide a visually grounded prompt~\cite{yang2023setofmark} and substantially improve the VLM's ability to identify and reason about the objects.

In addition to the VQA task, we evaluate the utility of MotionBits segmentation in downstream robotic manipulation tasks. For this evaluation, the same composite objects were compactly arranged on the tabletop, and the robot was tasked with stacking them into a tower using predicted segmentation masks for action planning. The evaluation protocol is detailed in \cref{sec:tower_stack_protocol}. Quantitatively, \cref{tab:towerstacktab} shows that baseline methods (SAM, SAMIV, and QwenVL) fail to provide reliable manipulation cues, whereas our MotionBits segmentations consistently yield the robust object localization required for task completion. Qualitatively, \cref{fig:application_study_visualizations} illustrates how SAM's over-segmentation leads to infeasible grasps and unstable placements. In contrast, our MotionBits segmentation masks, despite minor imperfections, significantly outperform baselines in stack height and success rate. These results demonstrate that motion-level segmentation from physical interactions effectively recovers underlying object geometries, enabling precise robot action planning for manipulating visually complex and unseen objects.

\begin{table}[t]
  \centering
  \caption{Results of the tower stacking task guided by object segmentation masks.}
  \label{tab:towerstacktab}
  \renewcommand{\arraystretch}{0.9} 
  \setlength{\tabcolsep}{7pt}      
  \setlength{\aboverulesep}{2pt}   
  \setlength{\belowrulesep}{2pt}   
  \scalebox{0.9}{%
  \begin{tabular}{@{}lccc@{}}
    \toprule
    Method & \# Stacked Objs & SR & Description of Failures \\
    \midrule
    SAM~\cite{Kirillov2023SegmentA} & 4 & 0 / 10 & Grasp \& Stack Failures \\
    SAMIV~\cite{seganymo} & 0 & 0 / 10 & No Segmentation \\
    QwenVL~\cite{Qwen2.5-VL} & 2 & 0 / 10 & Poor/No Segmentation \\
    \cmidrule(lr){1-4}
    \violetcell{\textbf{Ours}} & \violetcell{\textbf{37}} & \violetcell{\textbf{6 / 10}} & \violetcell{Cluttered Grasp Failures} \\
    \bottomrule
  \end{tabular}}
\end{table}

\section{Conclusion}
\label{sec:conclusion}
Understanding how rigid bodies interact is fundamental to embodied reasoning and robotic manipulation. However, most embodied systems remain limited by their vision modules due to reliance on semantic cues and neglect of real-world dynamics. In this work, we introduced a new concept defining the smallest unit in motion-based segmentation through kinematic spatial twist equivalence, enabling understanding of physical interactions. Our main contributions are threefold: (1) MotionBit: a segmentation unit where each rigid part exhibiting its own rigid-body motion is assigned its own mask, regardless of semantics, (2) MoRiBo: a hand-labeled benchmark for moving rigid-body segmentation across robotic manipulation and human-in-the-wild domains, and (3) a learning-free graph-based MotionBits segmentation method that achieves 37.3\% higher macro-averaged mIoU over state-of-the-art embodied perception methods. Beyond evaluation on the MoRiBo benchmark, we show that MotionBits segmentation masks provide actionable cues for embodied reasoning, enabling robots to interpret their environment and manipulate complex objects.

\newpage
{
    \bibliographystyle{unsrtnat}
    \bibliography{main}  
}

\clearpage
\appendix

\setcounter{section}{0}

\renewcommand{\thesection}{\Alph{section}}
\renewcommand{\thefigure}{S\arabic{figure}}
\renewcommand{\thetable}{S\arabic{table}}

\renewcommand{\thesection}{\Alph{section}}
\renewcommand{\theHsection}{Supp.\thesection}

\begin{center}
    {\LARGE \bf Supplementary Materials\par}
    \vspace{1em}
\end{center}

\section{Implementation Details}
In \cref{sec:method}, we introduced our learning-free graph-based MotionBits segmentation method. In this section, we provide the specific parameter values used in our implementation. All parameters were derived empirically for standard videos with resolutions of $480 \times 640$ or $640 \times 480$. Additionally, all matrix operations are computed using CUDA kernels for accelerated execution, and all experiments were conducted on a system equipped with a single NVIDIA RTX 4090 GPU.

\thirdorder{MotionBits Graph Construction}For constructing the spatial twist similarity graph described in \cref{sec:method_graph_construct}, we uniformly sample a grid of $\sqrt{n} \times \sqrt{n}$ nodes over the image. Through experimentation, we found that a grid of $100 \times 100$ ($n=10{,}000$ nodes) provides a strong balance between motion coverage and execution speed. When computing local spatial twists, we set $k=5$ to define each node's $k$-nearest neighbors. 

\begin{figure}[b]
    \centering
    \includegraphics[width=0.35\linewidth]{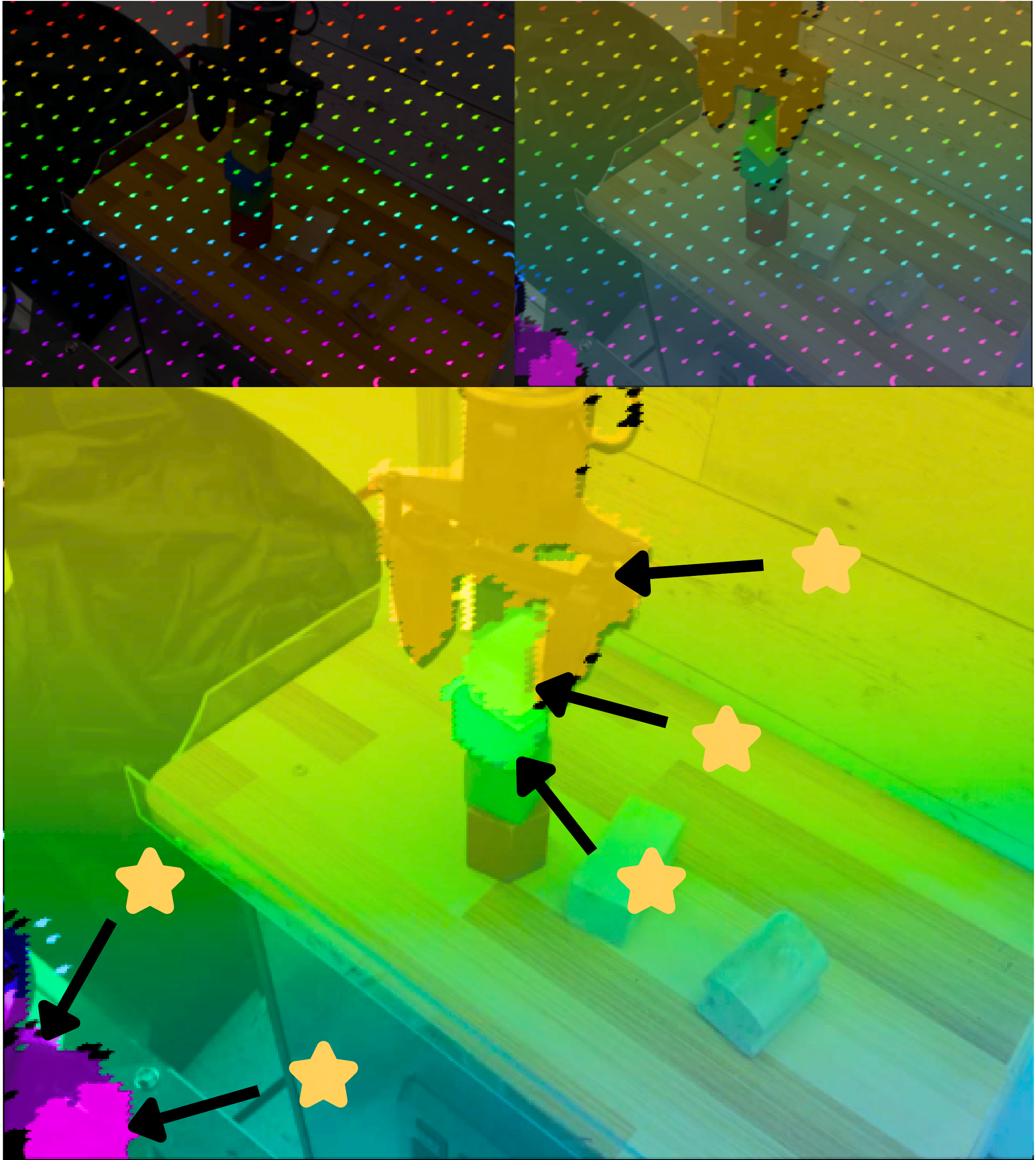}
    \caption{Iterative soft label propagation visualized (\cref{sec:method_soft_hard_seg} and line 8 of \cref{alg:motionbits_loop}). First, a set of seed nodes $S$ is uniformly selected from the set of graph nodes $\mathcal{M}$, and each seed node $s \in S$ is assigned a different ``colored paint''. Then, by continuously pouring the assigned color of paint onto each seed node and allowing the paints to flow and blend naturally according to the structure of graph $G$, we capture spatial twist motion regions through a color-coded soft MotionBits segmentation mask.}
    \label{supp_fig:softlabelprop_vis}
\end{figure}

\thirdorder{Soft and Hard MotionBits Segmentation}In \cref{sec:method_soft_hard_seg}, we create a global motion embedding by using the spatial twist similarity graph to iteratively diffuse ``paint'' through selected seed nodes. We empirically identified a seed count of $C=n*0.04$ (\textit{i.e.}, $C=400$ for $n=10{,}000$) as optimal for capturing meaningful motion regions without oversaturating the final embedding. Note that the granularity of MotionBits segmentation is directly affected by the parameters $n$ and $C$. If the grid resolution is too coarse, small motion regions may be treated as noise or skipped during local spatial twist estimation due to insufficient node coverage. A motion region may also be missed during soft label propagation if its nodes form a disconnected component and no seed is selected within that component. Increasing $n$ mitigates these issues but proportionally increases computation time. For the iterative process, we perform $R=100$ iterations of paint diffusion and inject additional paint every $r^*=5$ iterations. After completing soft label propagation, we discretize the global motion embedding into hard MotionBits segmentation using Markov clustering. This algorithm has two primary parameters, inflation and expansion, which we set to $1.15$ and $2$, respectively.

\section{Tower Stacking Evaluation Protocol}
\label[supp]{sec:tower_stack_protocol}
To evaluate the efficacy of MotionBits in real-world downstream robotic manipulation tasks, we conducted a tower stacking experiment using the five glued composite objects shown in \cref{fig:application_study_visualizations}. A fixed RGB camera is positioned approximately 75cm above the tabletop at a 60-degree tilt relative to the vertical axis. Each trial begins with a random robotic interaction (e.g. a push primitive) to generate motion cues. The resulting video is processed by the evaluated method to produce object segmentation masks, which inform a greedy stacking controller. This controller identifies the centroid and principal axes of the largest mask, determined by pixel area, to calculate the grasp point and orientation. This process is repeated until all identified objects are stacked or trial termination occurs. A trial is considered a success only if all five objects are successfully stacked into a stable tower. Any grasp or stack failure terminates the trial immediately. We conducted ten trials per method to ensure statistical significance.

\section{Additional Visualizations}
\thirdorder{Soft MotionBits Segmentation}We present additional visualizations of the soft label propagation process used to diffuse local motion affinities into a smooth global embedding (\cref{supp_fig:softlabelprop_vis}). Intuitively, differently colored ``paints'' flow naturally through the structure of the spatial twist similarity graph, which softly color-codes distinct MotionBit regions. 

\begin{figure}
    \centering
    \includegraphics[width=0.5\linewidth]{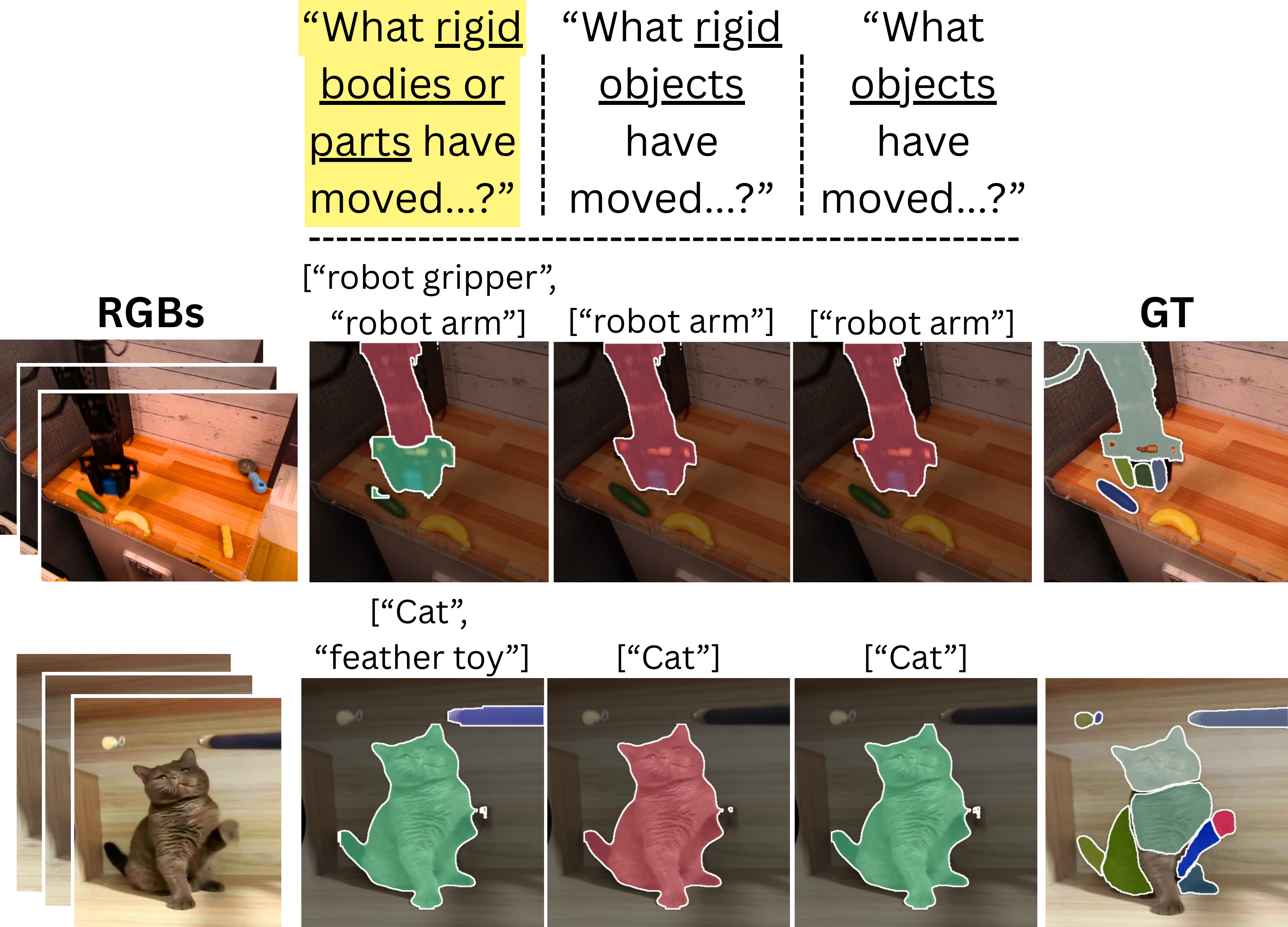}
    \caption{Examples of Qwen2.5VL-32B~\cite{Qwen2.5-VL} responses to different prompts. The prompt that empirically yielded the most accurate moving rigid-body segmentation results (highlighted) was used to evaluate VLMs on the MoRiBo benchmark, reflected in \cref{subsec:experiment_setup}.}
    \label{supp_fig:qwen_responses}
\end{figure}

\thirdorder{VLM Prompt Variations}We also provide comparisons between VLM prompt variations, from which we selected the best for our evaluations (\cref{supp_fig:qwen_responses}). 

\begin{figure}[b]
    \centering
    \includegraphics[width=0.85\linewidth]{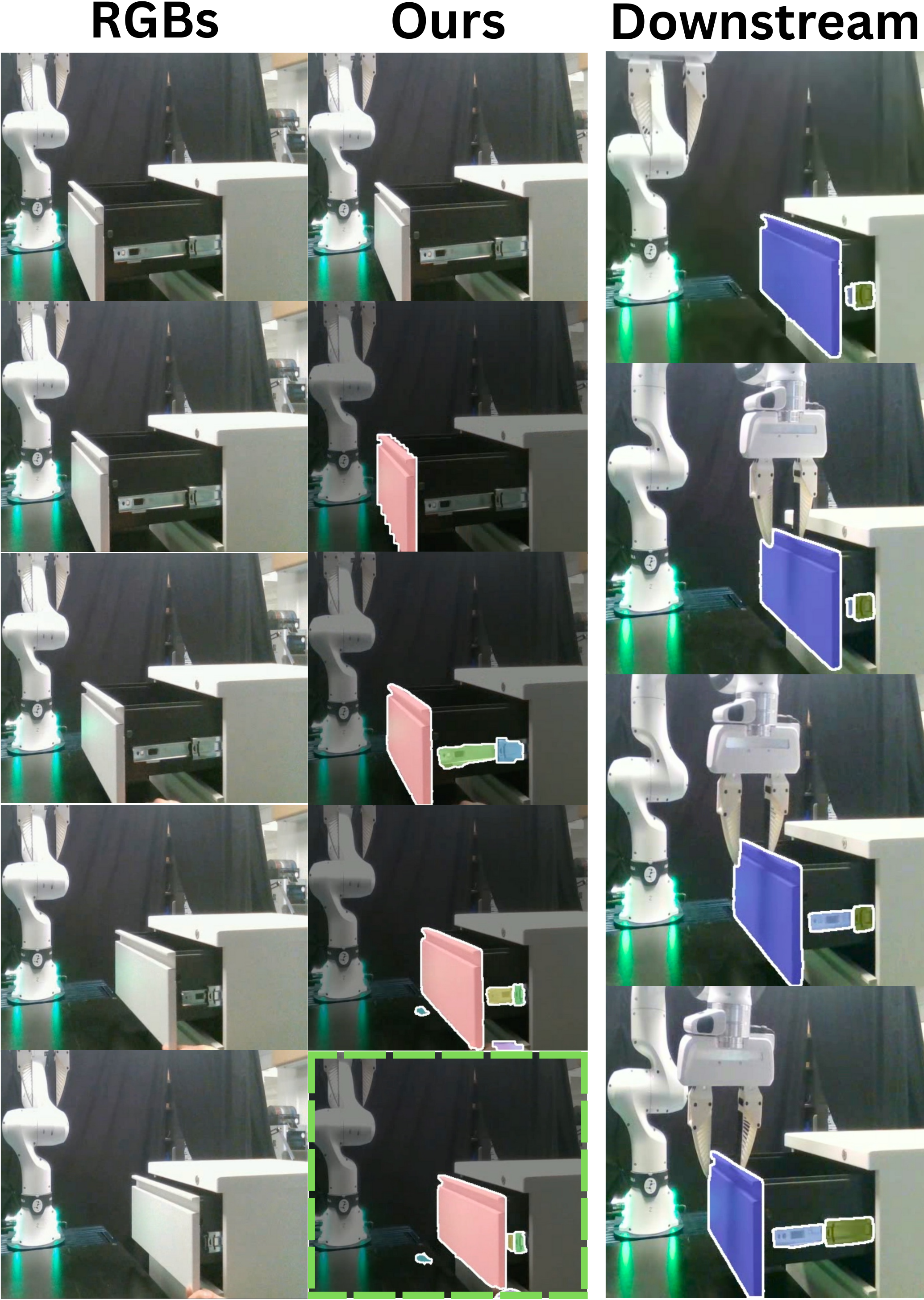}
    \caption{Additional example of MotionBits segmentation applied to the downstream task of opening a cabinet drawer. When a person interacts with the drawer, our method accurately segments the relevant MotionBits, including the drawer door and the two components of the drawer slider. This segmentation provides interaction-level cues that enable an embodied system to reason about and manipulate the cabinet.}
    \label{supp_fig:more_downstream_apps_1}
\end{figure}
\begin{figure}[b]
    \centering
    \includegraphics[width=0.85\linewidth]{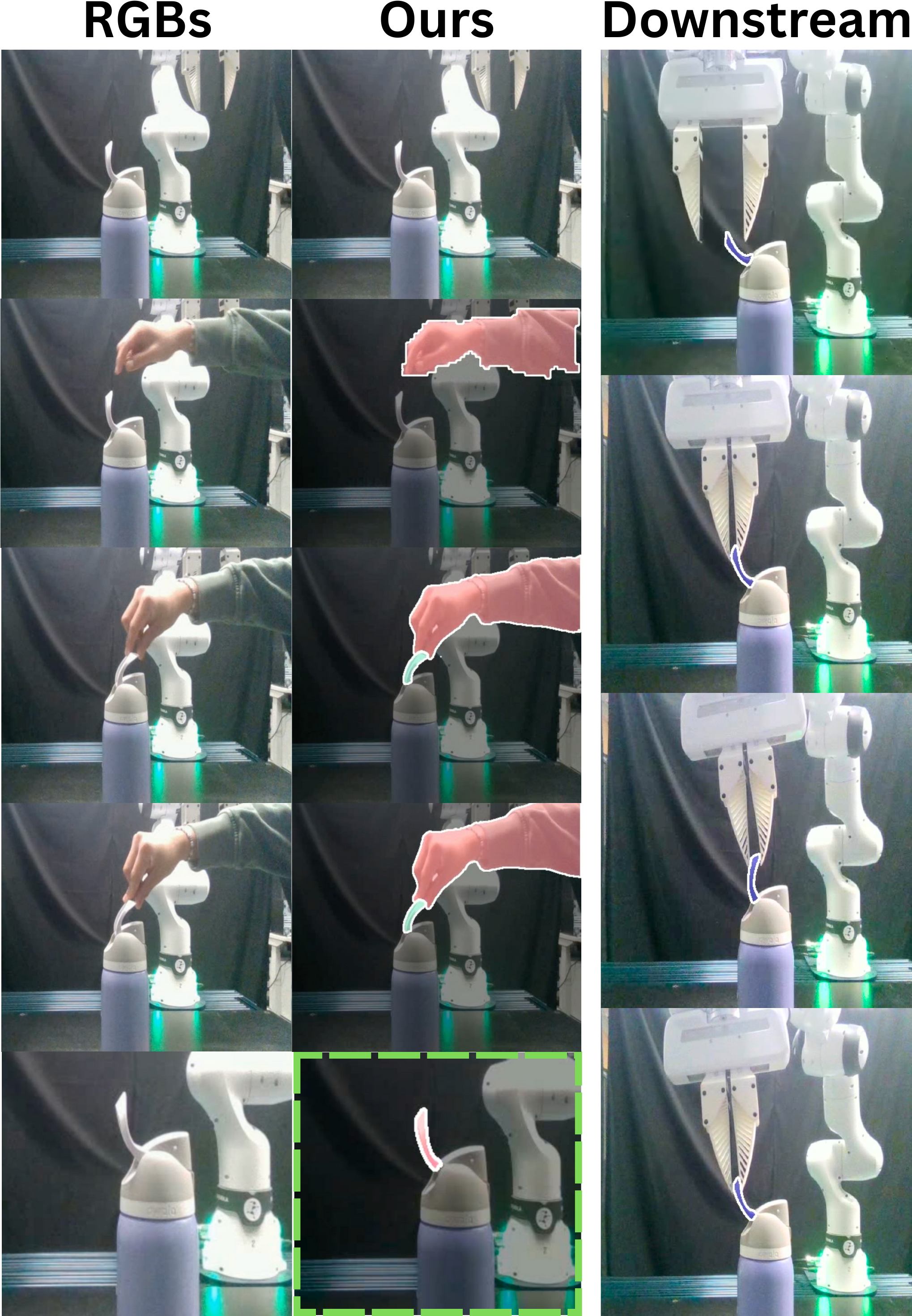}
    \caption{Additional example of MotionBits segmentation applied to the downstream task of manipulating the water bottle handle. When a person manipulates the water bottle handle, our method accurately segments the handle as a MotionBit, which provides interaction-level cues that enable a robot to manipulate the handle in a similar manner.}
    \label{supp_fig:more_downstream_apps_2}
\end{figure}

\thirdorder{Additional MotionBits Downstream Applications}Moreover, we provide two additional downstream examples beyond those shown in \cref{fig:application_study_visualizations}, demonstrating how our MotionBits segmentation supports embodied manipulation. In \cref{supp_fig:more_downstream_apps_1}, when a person interacts with the cabinet drawer, our method accurately segments the door and the two-part drawer slider as distinct MotionBits. This segmentation provides interaction-level cues that enable a robot to reason about and manipulate the cabinet. Similarly, in \cref{supp_fig:more_downstream_apps_2}, when a person interacts with the water bottle handle, our method accurately segments the moving handle, enabling a robot to manipulate it in an analogous manner.

\begin{figure*}[t]
    \centering
    \includegraphics[width=0.85\linewidth]{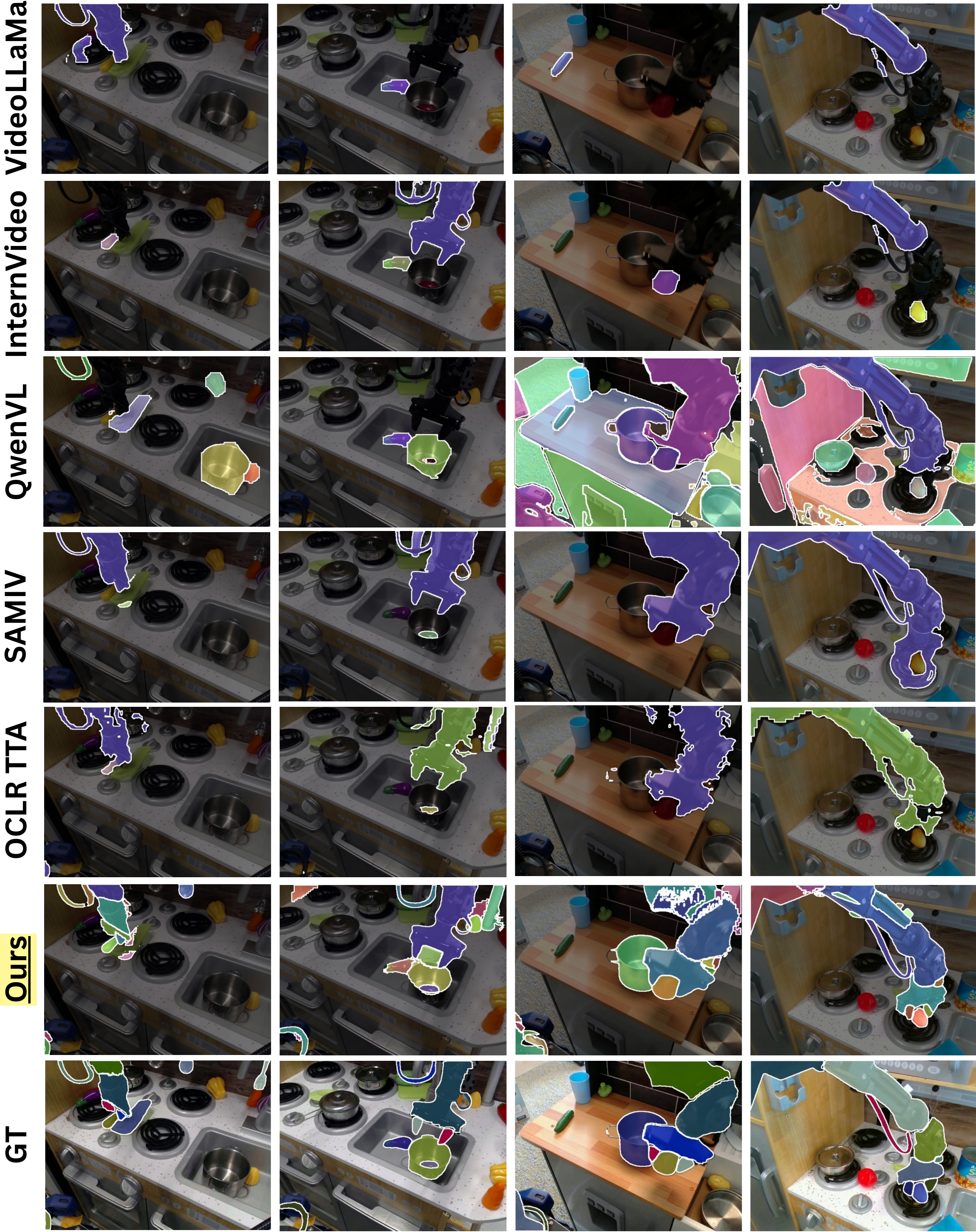}
    \caption{Extended qualitative comparisons of moving rigid-body segmentation on the Robotic Manipulation track of the MoRiBo benchmark between VideoLLaMa~\cite{damonlpsg2023videollama}, InternVideo~\cite{internvl}, QwenVL~\cite{Qwen2.5-VL}, SAMIV~\cite{seganymo}, OCLR TTA~\cite{xie2022segmenting} and our method. These results demonstrate that existing perception methods lack fundamental motion-level grounding, which is essential for embodied systems to reason about and interact with real-world environments. Our method is able to effectively identify moving rigid bodies within a scene.}
    \label{supp_fig:more_qualitative_results1}
\end{figure*}

\begin{figure*}[t]
    \centering
    \includegraphics[width=0.7\linewidth]{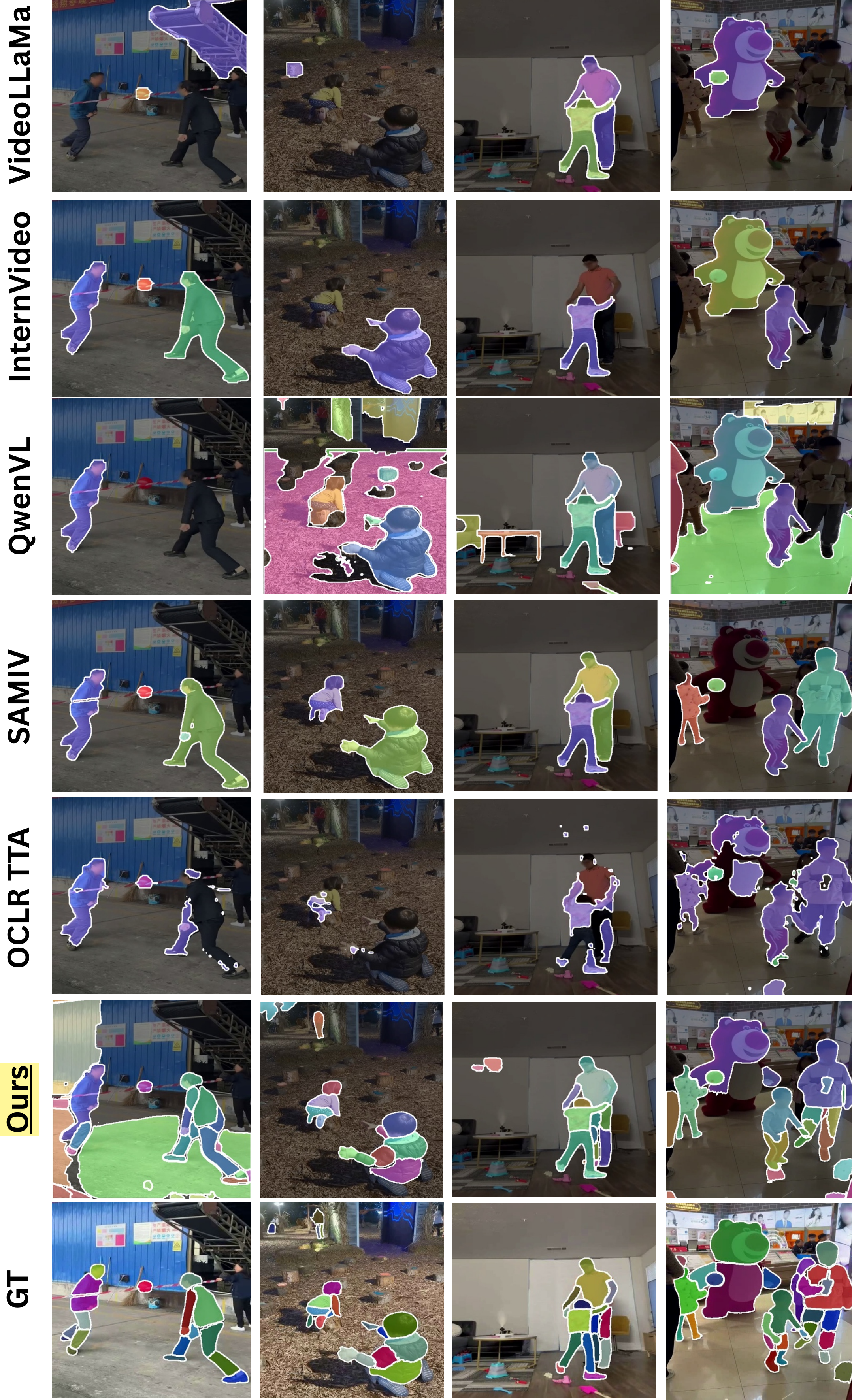}
    \caption{Extended qualitative comparisons of moving rigid-body segmentation on the Human-in-the-Wild track of the MoRiBo benchmark between VideoLLaMa~\cite{damonlpsg2023videollama}, InternVideo~\cite{internvl}, QwenVL~\cite{Qwen2.5-VL}, SAMIV~\cite{seganymo}, OCLR TTA~\cite{xie2022segmenting} and our method. These results demonstrate that existing perception methods lack fundamental motion-level grounding, which is essential for embodied systems to reason about and interact with real-world environments. Our method is able to effectively identify moving rigid bodies within a scene.}
    \label{supp_fig:more_qualitative_results2}
\end{figure*}

\thirdorder{Extended Qualitative Segmentation Comparisons}Furthermore, we provide extended qualitative comparisons between our method and baselines for moving rigid-body segmentation on the MoRiBo benchmark (\cref{supp_fig:more_qualitative_results1} and \cref{supp_fig:more_qualitative_results2}). These comparisons further showcase that existing embodied vision modules lack interaction-level perception, which is essential for downstream embodied tasks. 

\begin{figure*}[t]
    \centering
    \includegraphics[width=0.85\linewidth]{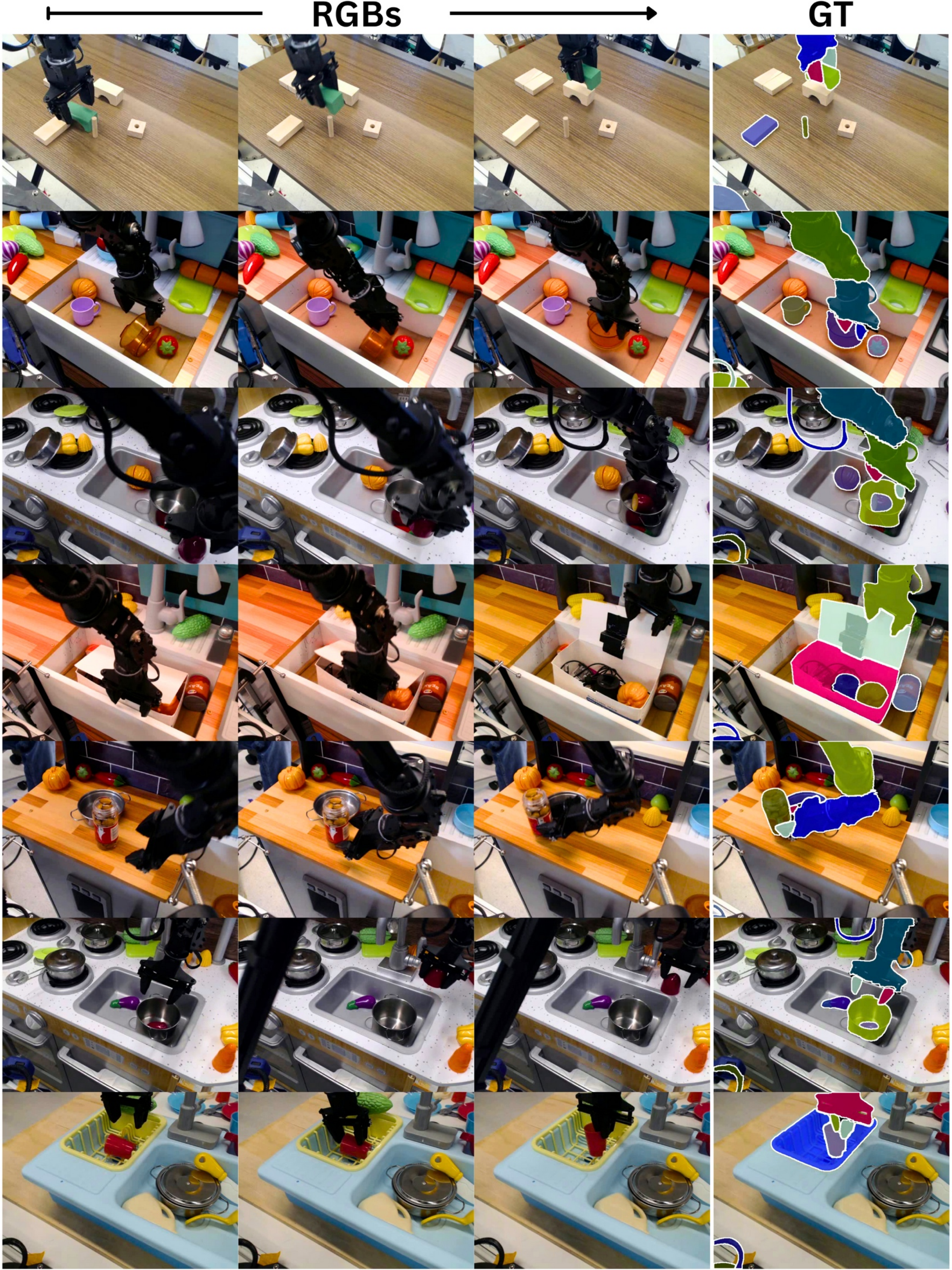}
    \caption{More examples from the Robotic Manipulation track of the proposed MoRiBo benchmark (\cref{sec:benchmark}).}
    \label{supp_fig:benchmark_ex1}
\end{figure*}

\begin{figure*}[t]
    \centering
    \includegraphics[width=0.85\linewidth]{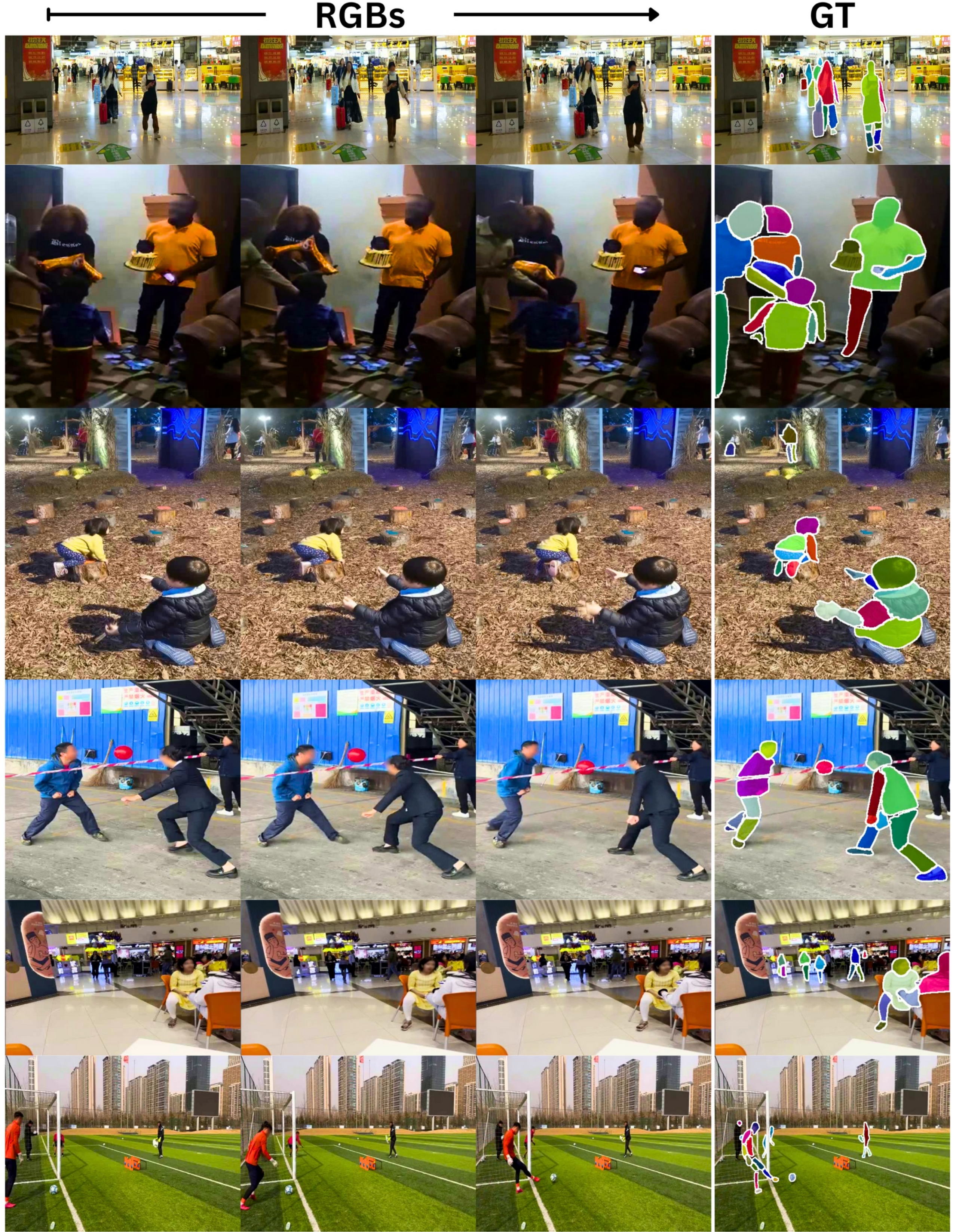}
    \caption{More examples from the Human-in-the-Wild track of the proposed MoRiBo benchmark (\cref{sec:benchmark}).}
    \label{supp_fig:benchmark_ex2}
\end{figure*}

\thirdorder{Extended MoRiBo Benchmark Examples}We also provide additional examples from both the Robotic Manipulation (\cref{supp_fig:benchmark_ex1}) and Human-in-the-Wild (\cref{supp_fig:benchmark_ex2}) tracks of the MoRiBo benchmark, highlighting the diverse scenarios it offers for evaluating both current and future methods.

\end{document}